\documentclass[nonacm,acmtog]{acmart}

\usepackage{booktabs} 
\usepackage{makecell}
\usepackage{subcaption}

\usepackage{svg}

\usepackage[ruled]{algorithm2e} 

\SetAlFnt{\small}
\SetAlCapFnt{\small}
\SetAlCapNameFnt{\small}
\SetAlCapHSkip{0pt}

\def\ifarXiv#1{#1}
\def\ifNotarXiv#1{}

\newcommand{\ignore}[1]{}

\DeclareMathAlphabet{\mathbfit}{OML}{cmm}{b}{it}

\makeatletter
\DeclareRobustCommand\onedot{\futurelet\@let@token\@onedot}
\def\@onedot{\ifx\@let@token.\else.\null\fi\xspace}

\makeatother

\definecolor{MyDarkBlue}{rgb}{0,0.08,1}
\definecolor{MyAqua}{rgb}{0,0.7,0.7}
\definecolor{MyDarkGreen}{rgb}{0.02,0.6,0.02}
\definecolor{MyDarkRed}{rgb}{0.8,0.02,0.02}
\definecolor{MyDarkOrange}{rgb}{0.40,0.2,0.02}
\definecolor{MyPurple}{RGB}{111,0,255}
\definecolor{MyRed}{rgb}{1.0,0.0,0.0}
\definecolor{MyGold}{rgb}{0.75,0.6,0.12}
\definecolor{MyDarkgray}{rgb}{0.66, 0.66, 0.66}

\newif\ifdrafting
\draftingtrue 
\ifdrafting

    \newcommand{\cih}[1]{{\textcolor{MyAqua}{[Charles: #1]}}}

\else
    \newcommand{\ds}[1]{}
    \newcommand{\cih}[1]{}
    \newcommand{\jw}[1]{}
    \newcommand{\ky}[1]{}
    \newcommand{\samirag}[1]{}
\fi

\SetKwFor{For}{for }{}{}

\newcommand{\myparagraph}[1]{\vspace{0.1cm}\noindent\textbf{#1}}

\usepackage[capitalize]{cleveref}
\usepackage{svg}
\crefname{section}{Sec.}{Secs.}
\Crefname{section}{Section}{Sections}
\Crefname{table}{Table}{Tables}
\crefname{table}{Tab.}{Tabs.}

\def\ourname{DreamWalk}
\def\figstylegridmacaron{13}
\def\figmixing{14}
\def\figdbxstyle{15}

\begin{document}

\title{DreamWalk: Style Space Exploration using Diffusion Guidance}

\author{Michelle Shu*\textsuperscript{1} \and 
Charles Herrmann*\textsuperscript{3} \and 
Richard S. Bowen\textsuperscript{2} \and 
Forrester Cole\textsuperscript{3} \and
Ramin Zabih\textsuperscript{1,3}  \\
\textsuperscript{1}Cornell University,
\textsuperscript{2}Geomagical Labs,
\textsuperscript{3}Google Research
}

\begin{abstract}

Text-conditioned diffusion models can generate impressive images, but fall short when it comes to fine-grained control. Unlike direct-editing tools like Photoshop, text conditioned models require the artist to perform ``prompt engineering,'' constructing special text sentences to control the style or amount of a particular subject present in the output image. Our goal is to provide fine-grained control over the style and substance specified by the prompt, for example to adjust the intensity of styles in different regions of the image (Figure~\ref{fig:teaser}). Our approach is to decompose the text prompt into conceptual elements, and apply a separate guidance term for each element in a single diffusion process. We introduce \emph{guidance scale functions} to control \emph{when} in the diffusion process and \emph{where} in the image to intervene. Since the method is based solely on adjusting diffusion guidance, it does not require fine-tuning or manipulating the internal layers of the diffusion model's neural network, and can be used in conjunction with LoRA- or DreamBooth-trained models (Figure~\ref{fig:dbteaser}). 
Project page: \url{https://mshu1.github.io/dreamwalk.github.io/}

\end{abstract}

\twocolumn[{%
\renewcommand\twocolumn[1][]{#1}%
\maketitle
\thispagestyle{empty}

\begin{center}
    \flushleft
    \captionsetup{type=figure}
    \includegraphics[width=\linewidth]{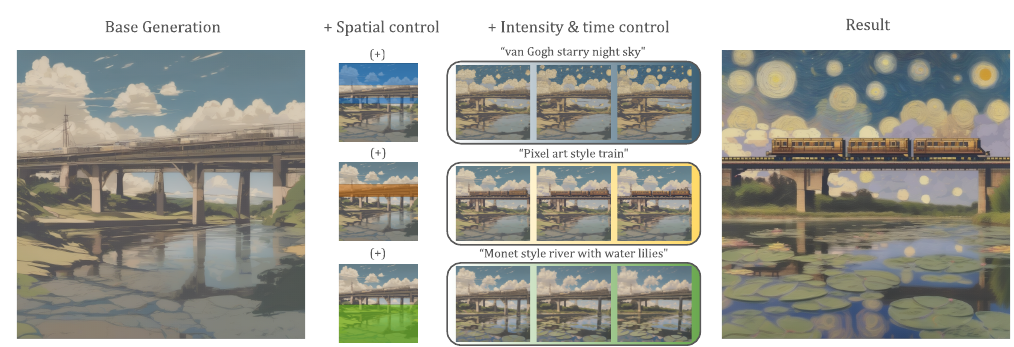}
    \captionof{figure}{
    DreamWalk allows fine-grained control of style text-to-image generation. We start with a base generated image (left), using the prompt ``A river flows under a bridge with clear sky''. 
    We explore style space by independently increasing different styles at different locations. 
    The middle row at center shows three images generated by increasing the pixel art style (using the prompt ``Pixel art style train''), applied to the orange mask shown in the (+) column.
    The image at right shows the result of increasing all three styles in their mask-specified locations. 
    All images shown are generated directly from diffusion using our guidance scale functions, and do not rely on image compositing or other post-processing. Generated with SDXL.
    }
    \label{fig:teaser}
\end{center}%
}]

\section{Introduction}

While the quality of text-to-image generation has dramatically improved in the last couple of years thanks to diffusion models, the best way to interact and control these methods still remains unclear. Many techniques center around Standard tools like Photoshop or Procreate can apply a wide range of effects, ranging from brushes that change the image in precise local ways, to image filters that globally alter the image. 
These effects come with a variety of user-adjustable parameters that provide fine-grained control. 
For example, a sepia filter can be applied with adjustable magnitude or tint.

Generative models offer new and powerful capabilities, and can create a completely new image from a simple text prompt such as ``a river flows under a bridge with clear sky''. 
For a given text prompt, sampling different seeds allows for exploring new compositions or scenes with elements from that prompt. 
Styles can be applied to these scenes by adding a style prompt -- ``a river flows under a bridge with clear sky in the style of van Gogh''. 
However, changes to the prompt often come with large changes in the generated image's composition or layout and it is difficult to use a prompt to granularly control the intensity or the location of an applied style.%

Our goal is to support user control over the amount, location, and type of style being applied, and thus use diffusion models to explore the style space for a given generated image. 

We generalize the multiple guidance formulation of \cite{liu2022compositional} in order to decompose prompts into different conceptual elements, such as a subject prompt and a style prompt; these can be granularly emphasized or de-emphasized. 
We introduce \emph{guidance scale functions}, which allow the user to control when in the diffusion process and where on the image each guidance term is applied. 
The control is \emph{fine-grained} in that the emphasis can be modified on a real-valued scale, which provides more control than just adding language to the prompt. 
It is \emph{flexible} in that emphasis can be applied spatially (to all or just part of the image) and temporally: emphasizing a prompt early in the diffusion process causes it to affect overall composition or layout more strongly, whereas applying it later causes it to affect texture or image details (see Section~\ref{sec:timesteps}).

Unlike many other approaches to control like ControlNet~\cite{zhang2023adding} or StyleDrop~\cite{sohn2023styledrop}, our approach does not require fine-tuning or training of any kind. 
We also do not access the internal components of the network like other techniques such as Prompt-to-Prompt~\cite{hertz2022prompt} or Gligen~\cite{li2023gligen} and as a result our technique does not need to be adapted or changed based on the architecture. 
Our method comes directly out of the diffusion formulation and as such can be applied to any diffusion network trained with text guidance~\cite{ho2022classifier}.

Our technique can control the application of style, as shown in Figure~\ref{fig:teaser}, as well as the degree of personalization shown in Figure~\ref{fig:dbteaser}. 
We provide results on both Stable Diffusion 1.5 and Stable Diffusion XL, two diffusion models with different architectures, and demonstrate that our technique works with both standard text-to-image models as well as DreamBooth and LORA fine-tunings.
\ifNotarXiv{
Additional examples and baselines are provided in the supplemental material.
}

\begin{figure}
\includegraphics[width=0.45\textwidth]{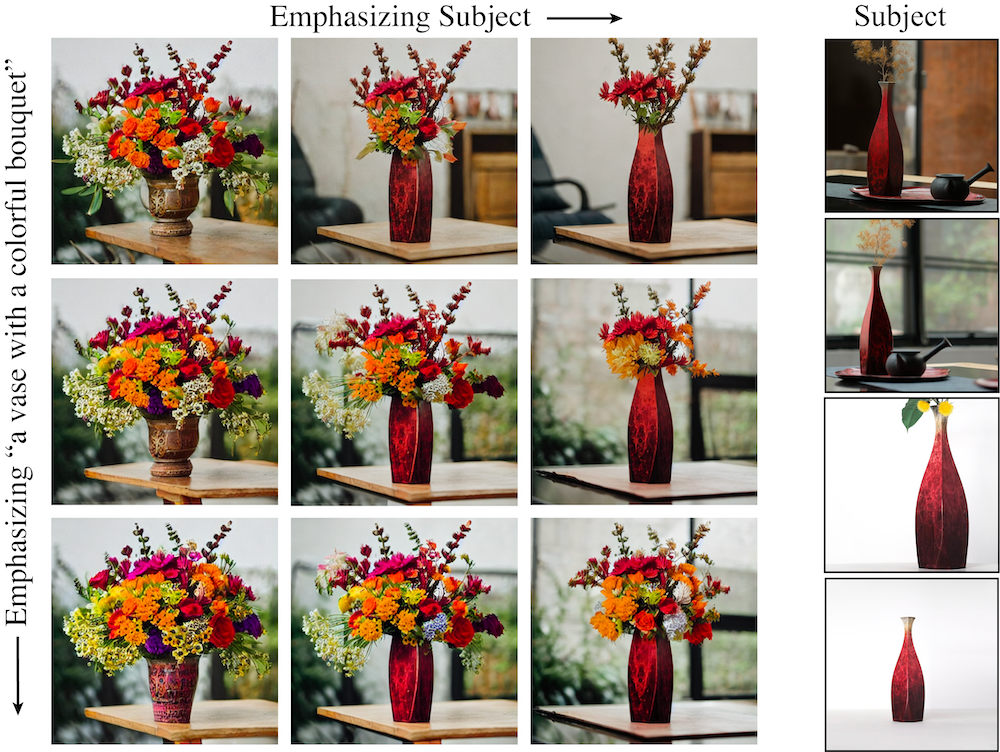}
\caption{\textbf{Controllable subject / prompt emphasis.} Our formulation can explore adherence to a DreamBooth subject or adherence to the text prompt. Generated with SD1.5.}
\label{fig:dbteaser}
\end{figure}

\section{Background and Related Work}
\subsection{Diffusion Models and Guidance Functions}\label{sec:diffusion}
Denoising diffusion models\cite{sohl2015deep,ho2020denoising,Yang2022DiffusionMA} are a class of generative models which are trained to transform a sample of Gaussian noise into a sample of a data distribution. 
To generate a sample from the data distribution, we start with $x_1 \sim N(0, I)$ as Gaussian noise. For a given time step $t$ from $1$ to $0$, we alternate between denoising  and adding back a smaller amount of Gaussian noise. Over the course of the sampling process, from $t=1$ to $t=0$, the image becomes less noisy and more realistic. An update step using a DDPM~\cite{ho2020denoising} sampler might be:

\begin{equation}
x_{t-1} = (x_t -  f_\theta(x_t, t)) + \epsilon_t\label{eq:ddpm}
\end{equation}

where $f_\theta(x_t, t)$ is the guidance function, which is responsible for the denoising stage of the diffusion process. The guidance function may combine multiple forward passes of the diffusion model, $f_\theta(x_t, t, c)$, a network trained to estimate the noise of the image given a condition $c$. For example, classifier-free guidance~\cite{ho2022classifier} combines conditional and unconditional outputs to improve adherence to the conditioning vector.

\begin{equation}
f_\theta(x_t, t) = f_\theta(x_t, t, \emptyset) +  s_i \cdot \left[f_\theta(x_t, t, c_i) - f_\theta(x_t, t, \emptyset)\right]
\end{equation}

Liu, et al.~\cite{liu2022compositional} demonstrated that multiple diffusion forward passes may be composed to to allow the image generator to ``build'' a scene from various elements. 
For example, they combine phrases like ``a photo of cherry blossom trees'', ``sun dog'', and ``green grass'' in order to generate an image with components from each of these phrases. 
Given a series of conditions $c_0, \ldots, c_k$ defined on a single model $\theta$, they compute a single overall update step as follows:

\begin{equation}
f_\theta(x_t, t) = f_\theta(x_t, t, \emptyset) + \sum_{i=1,\ldots,k}  s_i \cdot \left[f_\theta(x_t, t, c_i) - f_\theta(x_t, t, \emptyset)\right]
\label{eq:compose}
\end{equation}

where $s_i$ are the guidance scales or coefficients (see Equation 15 in \cite{liu2022compositional}). Note that if $k=1$, $c_0$ is a text-conditioning vector and $\emptyset$ is an empty conditioning vector (``null text''), Eq.~\ref{eq:compose} is equivalent to classifier-free guidance~\cite{ho2022classifier}.  

We build upon \cite{liu2022compositional}, while addressing a different task and providing a more general formulation. \cite{liu2022compositional} is focused on compositionality, with the goal of producing a single sample with all the elements present in the multiple text prompts (e.g. a tree, a road, a blue sky). In an image they produce, a particular element in the text is either present or absent; a success is defined as having all the elements in the multiple text prompts present in the generated image. In contrast to their true/false task, we focus on fine-grained control for stylization and personalization, producing multiple samples which offer a choice of target distributions.
To solve this very different task requires a generalization of the multiple guidance framework of \cite{liu2022compositional} through the introduction of guidance scale functions; this is described in section~\ref{sec:method} where the relationship with \cite{liu2022compositional} is discussed in more detail.

Multidiffusion~\cite{bar2023multidiffusion} and SyncDiffusion~\cite{lee2023syncdiffusion} use a similar formulation to create panoramas or images at a higher resolution than can be naturally generated by the diffusion model, which typically has a fixed generation resolution. Both methods do this by partitioning an image into overlapping windows and then running a single diffusion process where each window provides guidance to pixels in its regions. In both techniques, all the pixels go through a single, joint diffusion process. 

\subsection{Style for Image Creation}\label{sec:rwstyle}
Stylization or style transfer, taking a real photo and applying a style, is a long explored field, with early techniques involving hand designed algorithms designed to mimic a certain artistic style~\cite{haeberli1990paint, salisbury2023interactive, litwinowicz1997processing, hertzmann2003survey}, and parametric~\cite{heeger1995pyramid,portilla2000parametric} or non-parameteric~\cite{efros1999texture, barnes2009patchmatch, hertzmann2023image} techniques for matching image statistics to either new texture patches or stylized images. Recently, as deep learning has become more popular, the field has shifted towards neural style transfer and texture synthesis~\cite{gatys2015texture, jing2019neural, azadi2018multi}. Unlike these techniques, we do not aim to transfer image statistics from a style exemplar image. Instead, we focus on gaining fine-grained control over a generative model's pre-existing ability to apply style and move to between different styles.

\cite{kerras2019style} showed that modifying the generator architecture of a GAN can induce a latent space with coarse and fine features disentangled. A number of publications followed,  focusing on controllability by disentangling the latent space, producing interpretable paths or directions through the latent space that allow for continuous control of pose, zoom, scene composition, hair length, and many others: \cite{wu2021stylespace, tewari2020stylerig, voynov2020unsupervised, harkonen2020ganspace, shen2020interpreting}. Other methods tried to identify existing interpretable paths in the model ~\cite{jahanian2019steerability, harkonen2020ganspace, voynov2020unsupervised}.  \cite{patashnik2021styleclip} learned to use CLIP embeddings to control StyleGAN latents with text prompts. More recently, similar techniques have been applied to diffusion models: \cite{park2023understanding} introduces a pullback metric and uses this to find latent-space basis vectors with particular semantics; \cite{preechakul2022diffusion} learn to encode directly into the diffusion model's latent space, allowing interpolation along axes like age and hair color. \cite{wu2023uncovering} use a soft weighting of base and attribute prompts, with the weighting factor chosen to try to limit the impact of the attribute prompt only to appearance, and not the layout, of the generated image. This weighting factor is learned through an optimization problem and is done per image, though the weights can be applied to any image. Their main focus is taking a generated image and then applying the style once, without fine-grained control over how much the style is applied.

\subsection{Personalized Text-to-Image Generation}

Given the difficulty of capturing fine-grained concepts using text, considerable attention has been given to methods for personalizing image generation models. 

One approach is to train a new model that is conditioned on both images and text~\cite{chen2023suti,ma2023subject,shi2023instantbooth}, which requires gathering a large dataset and retraining. An alternative approach is to finetune an existing model, either by finetuning the text embeddings~\cite{gal2022image}, the weights of the generator network~\cite{ruiz2023dreambooth,sohn2023styledrop,kumari2023multi}, or both~\cite{avrahami2023break}. Most commonly, personalized finetuning takes 3-5 images of a specific object or style and finetunes the entire text-to-image network to reconstruct these images when conditioned with text of a rare token $[\mathcal{V}]$, placed in context of its noun, e.g.  ``a photo of $[\mathcal{V}]$ dog''. DreamBooth~\cite{ruiz2023dreambooth} also uses a class-preservation prior in order to prevent the noun's conditioning from drifting away from its original meaning and collapsing to only representing the 3-5 images being used for fine-tuning. 

Parameter-efficient finetuning~\cite{hu2021lora,houlsby2019parameter} such as LoRA is commonly used to speed up finetuning and reduce the size of the weight modifications, allowing users to transfer only the update weights instead of the entire network weights.

\section{Our Approach}
\label{sec:method}

Our overall goal is to be able to  ``walk'' or explore style space -- providing fine-grained control over different aspects of style, such as location, intensity, and type. To do this, we need three things:  (1) the ability to aim for a specific target distribution based on multiple conditions (Sections \ref{sec:multiple-guidance} and \ref{sec:controllable-walks}); (2) a way to create the appropriate conditions (Section \ref{sec:deconstructingprompts}); and (3) a mechanism to provide  fine-grain control over how we guide towards these conditions and their relative location, intensity, and type (Section \ref{sec:guidance_scale_functions}).

Our full framework allows for the robust application of various style. In Figures~\ref{fig:singlestylemanyapplications1} and \ref{fig:singlestylemanyapplications2}, we show a simple application of the framework, applying a single style to a given prompt. In later sections, we will demonstrate more complex capabilities, including controlling how style is applied in the image formation process and walking in style space.

\subsection{Multiple Guidance Formulation}
\label{sec:multiple-guidance}

Building on the multiple guidance terms introduced by \cite{liu2022compositional}, we propose \emph{guidance scale functions} $s_i(t, u, v)$ which allow for granular control over when, in the diffusion noise schedule, and where, in the image space, a forward pass with a specific condition is used. Specifically, we consider the following formulation for composition of diffusion models:

\begin{gather*}
f_\theta(x_t, t) = f_\theta(x_t, t, \emptyset) + \sum_{i=1,\ldots,k} s_{i}(t, u, v) g_\theta(x_t, t, c_i) \\
g_\theta(x_t, t, c_i) =  f_\theta(x_t, t, c_i) - f_\theta(x_t, t, \emptyset)
\label{eq:ourmethod}
\end{gather*}

where for simplicity we define $g_\theta(x_t, t, c_i)$ as the guidance term for condition $c_i$. The guidance term includes the forward passes of the diffusion model $f_\theta(x_t, t, c_i)$ and $f_\theta(x_t, t, \emptyset)$ and indicates a direction and magnitude towards a condition. This magnitude can be emphasized or de-emphasized by the guidance scale function, $s_i(t, u, v)$ (dependent on time $t$ and spatial location $u$ and $v$).

\subsection{Creating multiple guidance terms from text prompts}\label{sec:deconstructingprompts}
Now that we have the ability to target a specific distribution based on multiple conditions, we need a way to create the appropriate conditions for our task of controlling stylization and personalization.

\myparagraph{Decomposing prompts for stylization}
Numerous works~\cite{haeberli1990paint, salisbury2023interactive, hertzmann2003survey, rashtchian2023substance, wu2023not} consider the style and substance of an image to be two separate decomposable concepts: e.g., any subject can be painted in watercolor or created as a woodblock print. As a result, we explore decomposing the prompt into several constituent parts, primarily focusing on a ``base'' prompt consisting of a description of the scene (subject, action, setting, etc.) and a style component often indicated by an art medium, artistic period, or an artist (watercolor, Impressionism, Hokusai, etc.). For example, the prompt ``a river flows under a bridge with a clear sky in the style of van Gogh'' might decompose into a base ``a river flows under a bridge with a clear sky'' and style ``van Gogh''.

\myparagraph{Separating personalized subjects from their context.}
The same technique can be used to for subject personalization~\cite{ruiz2023dreambooth}. DreamBooth introduces a rare token as an adjective that modifies an existing subject, e.g. ``a $[\mathcal{V}]$ vase''. This adjective is then trained to produce a specific vase. The phrase is then placed in context of a complete text prompt, often containing other adjectives, verbs (actions), or other context (background, style, etc.), e.g. ``a $[\mathcal{V}]$ vase with a colorful bouquet''. This complete sentence is then used as the single text conditioning. 

In the multiple guidance formulation, this sentence can be decomposed into its component parts and then guided towards each of these distributions separately allowing for control over which of these relative concepts is the most important to the user. If there is a desired subject (``a $[\mathcal{V}]$ vase'') and a desired state (``with a colorful bouquet''), the user can decide which of these to emphasize. Figure \ref{fig:dbteaser} show examples allowing adherence to either the text prompt (background, action) or the subject (the DreamBooth token and subject). This control is useful since one common failure is the entanglement between a subject's pose / actions and the subject's appearance. 
If the images used for personalization share a similar pose, e.g. all images of a dog are front-facing and with the dog sitting, then the generator can struggle to change the pose of the dog. Our method can allow the user to emphasize the action separately from the token reducing this bias.

\begin{figure}[htb!]
\includegraphics[width=\linewidth,trim={0 40 0 10}]{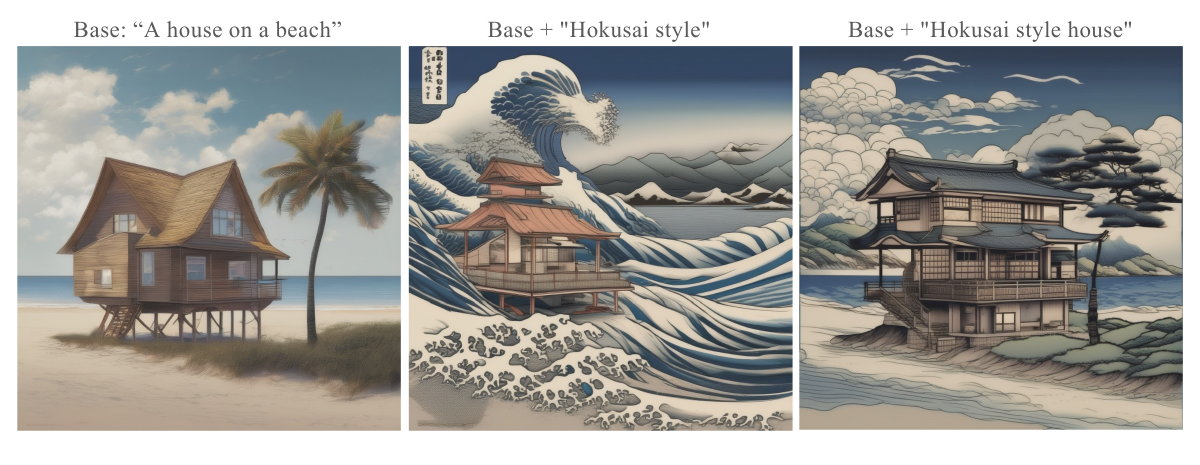}
\caption{Applying a style without specifying a subject can, depending on the biases of the prompt's distribution, can lead to changes in the subjects in the generated image. E.g., ``Hokusai style'' distribution has a strong bias towards showing Hokusai's famous waves even when they overwhelm the original content of the image. Adding a subject to the prompt, like ``house'', can mitigate this. Generated with SDXL.}
\label{fig:prompteffectonstyle}
\end{figure}

\subsubsection{Mitigating subject drift through prompting.} While these decompositions make sense in theory, in practice we observe that guiding towards a prompt which only contains a style phrase ``Hokusai style'' can induce unexpected results, as seen in Figure~\ref{fig:prompteffectonstyle}. Since text-to-image generative networks have been trained with pairs of images and text, style and substance have not been totally disentangled. This can produce ``subject drift'': while the user may like their content and just want it to look more like woodblock print, introducing ``Hokusai'' may also drive the content towards subjects that Hokusai depicted such as Mount Fuji, ocean waves, or Japanese style houses.

In practice, as shown in Figure \ref{fig:prompteffectonstyle}, we find that this can be mitigated by introducing a small amount of the subject matter into the style prompt. While this prompt engineering technique prevents wide-spread subject drift, there are smaller drifts that the user may or may not want; e.g. when applying ``Hokusai style'' to a building, the generation tends to produce Japanese style houses. In later sections, we show that this effect can be mitigated if desired through use of temporally-varying guidance scale functions.

\begin{figure*}[htb!]
\includegraphics[width=0.9\textwidth]{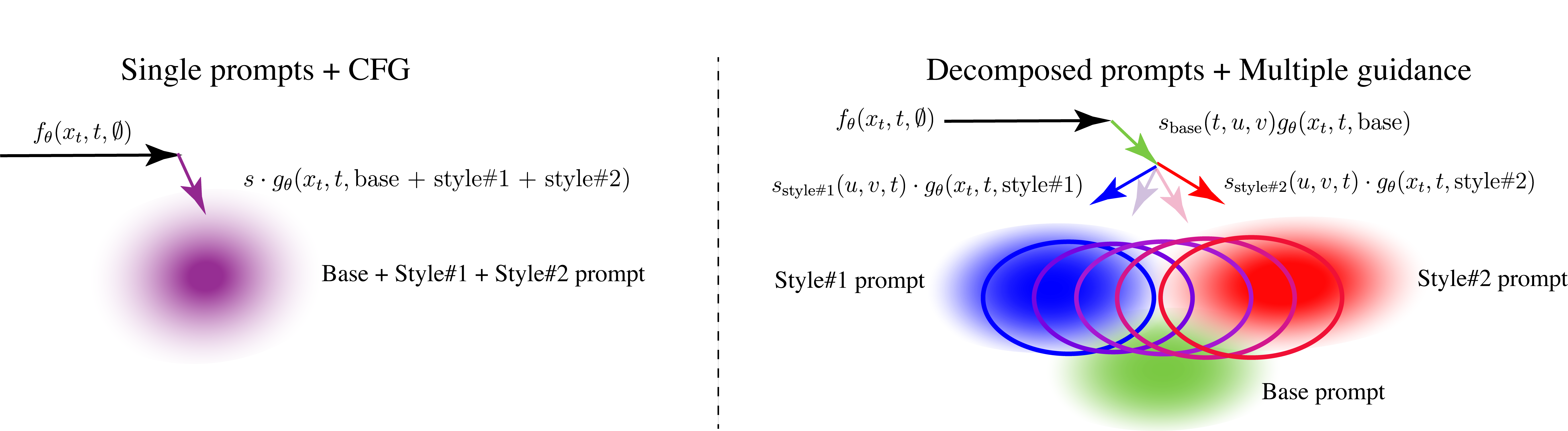}
\vspace{-2mm}
\caption{\textbf{Controlling the emphasis.} Left, the standard way of using text-to-image models is to specify everything in a single prompt and then guide towards it with the guidance term $g_\theta$ multiplied by scale $s$. However, this does not provide a natural way to emphasize individual parts of a single prompt. We note that prompts can be naturally decomposed, the simplest of which is base prompt (the objects, verbs, and setting; ``a river flows under a bridge with clear sky'') and any style application (``van Gogh''). Right, each decomposed prompt receives its own guidance term and scale function. By varying the scale function, we can walk towards a linear combination of terms, as denoted by the transparent arrows. This allows us to aim for any of the distributions represented by circle. This example depicts style interpolation with a layout from a base prompt. Note the unconditional distribution is not shown above.}
\label{fig:decompose_idea}
\end{figure*}

\subsection{Controllable walks using the guidance scale}
\label{sec:controllable-walks}

As described in \cite{ho2022classifier, liu2022compositional}, each guidance term pushes the final result towards its own distribution in the image manifold, with the the magnitude of the guidance scale determining how strong the push is. With a single conditional, this term is thought of as simply controlling how closely the image adheres to the single text prompt versus a generic image. With the same seed, a different magnitude would generate similar images that increasing adhere to the prompt (left of Fig.~\ref{fig:decompose_idea}).

This same concept applies in the multiple guidance formulation, with each guidance term's scale controlling how much the generated image adheres to that condition. 
In the compositionality problem that \cite{liu2022compositional} addresses, the goal is to generate a successful sample, defined as images where all the text elements are present.
In contrast, our goal is to sample from a user-controlled choice of distributions, selected to explore style space. 
Since we can control the guidance scale differently for each condition, this provides a choice of distributions to target as shown on the right of Figure~\ref{fig:decompose_idea}. 
For example, by choosing to target different distributions our formulation can walk between two different prompts (e.g. ``base'' and ``style''), interpolating between them, moving from a style focused distribution (red in the figure) to a base focused distribution (blue). Interpolating between multiple style distributions produces a style interpolation; emphasizing both produces a mixture. The scales between a base and style distribution provide control over how text adherence versus style. The addition of time-varying GSFs enables the base to control the layout of the final generation (see Sec~\ref{sec:guidance_scale_functions}).  Time-varying GSFs applied to DB tokens versus text allow us to control the importance of subject personalization versus text adherence as shown in Fig.~\ref{fig:dbteaser}.

\subsection{Using GSFs for fine-grained style control}\label{sec:guidance_scale_functions}

Once the prompts are decomposed as above, each guidance scale function controls the magnitude of the diffusion step for a single component. Critically, this modulation can depend on location within the image or on timestep. Here we show examples of using GSFs to control how a style applied to an image.

\begin{figure}
\newcommand{\normwidth}{0.23\linewidth}
\begin{tabular}{cccc}
\includegraphics[width=\normwidth]{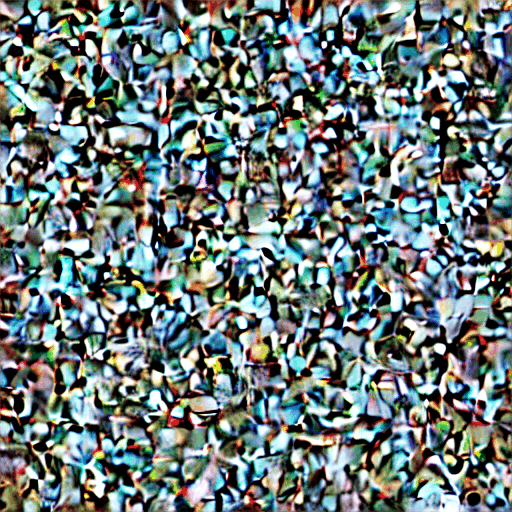}&
\includegraphics[width=\normwidth]{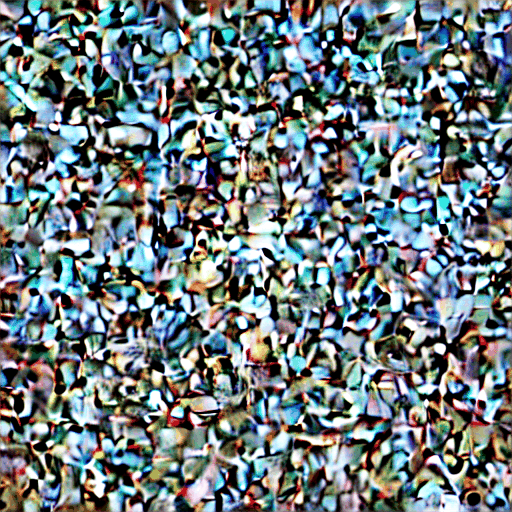}&
\includegraphics[width=\normwidth]{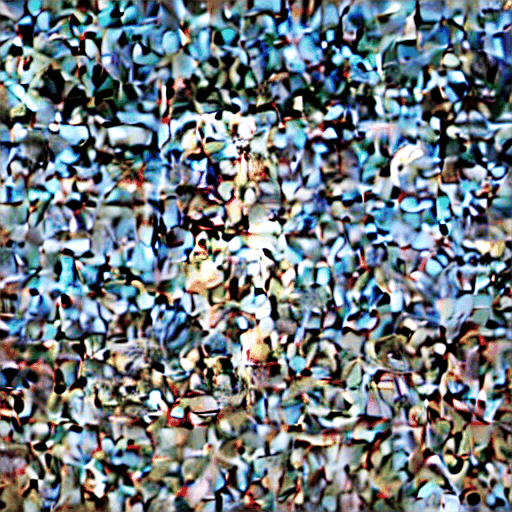}&
\includegraphics[width=\normwidth]{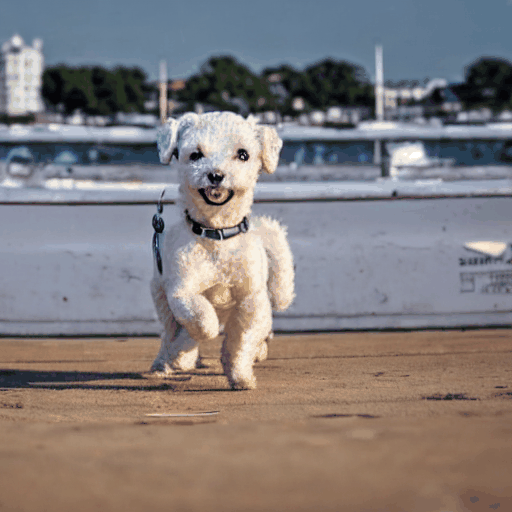} \\
\includegraphics[width=\normwidth]{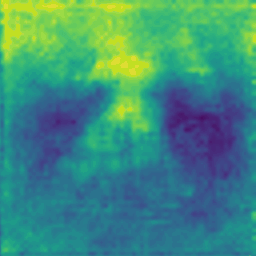}&
\includegraphics[width=\normwidth]{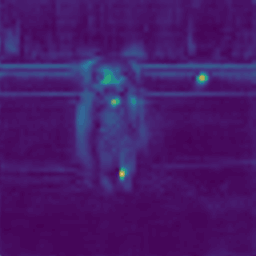}&
\includegraphics[width=\normwidth]{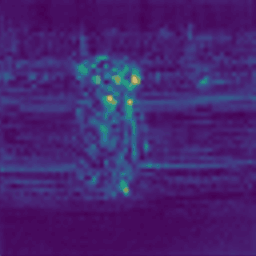}&
\includegraphics[width=\normwidth]{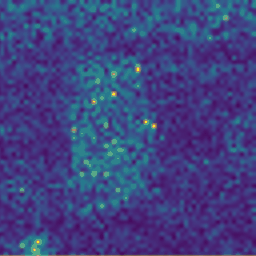}\\
$t=1$ & $t=0.8$ & $t=0.6$ & $t=0$
\end{tabular}
\caption{Examining the norm of the denoising predictions at different time-steps $t$ suggest that the image is formed coarse-to-fine. Near $t=1$, the network edits the global image layout (low frequencies) whereas near $t=0$ the network seems to focus on texture (high frequencies). Note, the top row is the latent at different $t$ steps decoded through the VAE, bottom row is norm of denoising prediction for text conditioned pass of the denoising network.}
\label{fig:image_formation}
\end{figure}

\subsubsection{Timestep dependence: disentangling style from content} \label{sec:timesteps}

\label{sec:texture_vs_geometry} As noted by several prior works including ~\cite{rissanen2022generative, meng2021sdedit}, the progressive denoising approach of diffusion models means that they implicitly create images in a coarse-to-fine manner. While this analysis does not directly apply to Stable Diffusion~\cite{rombach2022high} since it is a latent diffusion model which operates on the latent space (as opposed to pixel space), we find empirically that Stable Diffusion does form images from coarse to fine. Fig. \ref{fig:image_formation} shows the norm of the guidance terms, $\lVert g_\theta(x_t, t, \textrm{text})\rVert_2$, as a heat map, indicating which parts of the image are being edited at different timesteps. We observe that at $t=1$ the entire image is edited; at $t=0.8$, some of the edges and shapes are edited; and at $t=0$ only the high frequency textures are edited. This can be roughly observed in the decoded latent as well -- first the rough location of the dog is visible, and later details are filled in.

\begin{figure*}[htb!]
    \includegraphics[width=\linewidth]{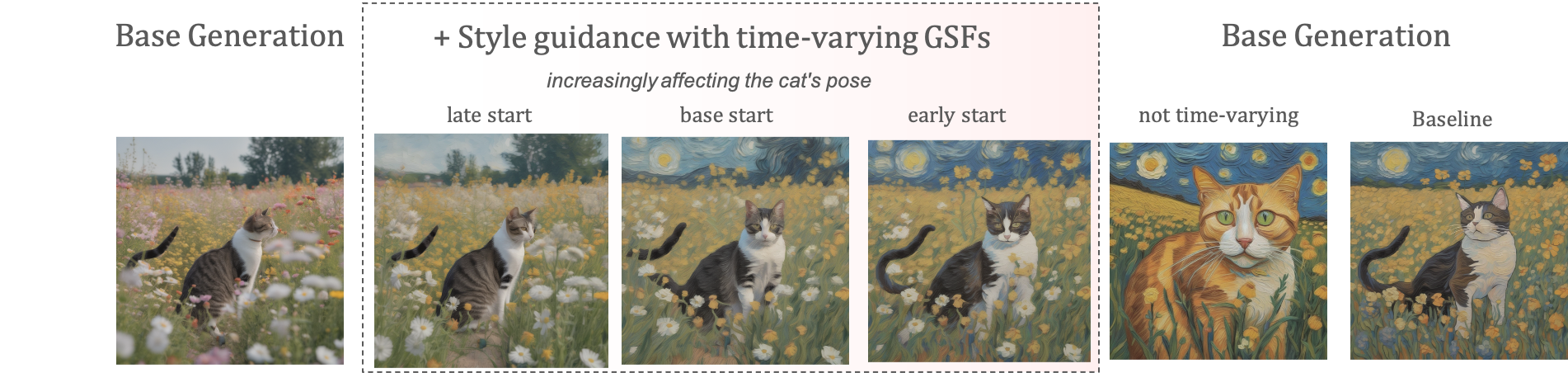}
    \caption{\textbf{Temporally-varying guidance scale functions.} Using different temporally-varying GSFs for the same style guidance, ``van Gogh'' leads to different effects. A higher guidance scale at $t=1$ leads to layout changes to the Base generation; roughly speaking, this acts as a knob that allows the user to control how much they want to allow style to affect the overall structure and composition of the image. Specifically, ``late start'' applies higher frequency components of a style, whereas ``early start'' changes both low and high frequency components. Base prompt: ``A cat walking in a field of flowers.'' Exact functions used are in the supplemental. Generated with SDXL.}
    \label{fig:timevary_types}
\end{figure*}

This observation motivates the technique shown in Figure~\ref{fig:timevary_types}. The base generation is ``a cat in a field of flowers''. Typically, applying a van Gogh style results in a change to the coarse layout of the generated image. This is true either when editing the prompt or when decomposing the prompt and applying a uniform guidance scale to the `van Gogh` style. Our general goal is to explore granular application of style while keeping the rough layout of the base generated image -- put simply, ``this composition, but as though van Gogh painted it''. Our observations regarding coarse-to-fine image generation directly motivate our proposed solution. Early in generation, we guide the image primarily towards the base prompt with little to no emphasis on the style prompt; later in generation, we guide the image primarily towards the style prompt with little to no emphasis on the base prompt. This results in a very similar composition to the base but with van Gogh's signature swirly paint. 

Technically for style, we propose a simple parameterized function which linearly increases and discuss the effects of different parameter settings of the functions.

$$s_{\textrm{style}}(t, u, v) = \max(0, \frac{m}{a}(a-t))$$

where conceptually $m$ is the magnitude of the scale and $a$ determines when the guidance term becomes non-zero. Since $t$ typically starts at $1$, if $a>1$ then the guidance happens at all stages of the diffusion process and this guidance term will have an ``early start'' non-trivially affecting the layout. If $a<1$, then the guidance term becomes non-zero later in the diffusion process and the guidance term has a ``late start'' and does not affect layout or medium frequency elements. ``base start'' is $a=1$ and acts as an intermediate or default option for this type of time-varying. For the base guidance term, we simply use a linearly decreasing GSF, $s_i(t) = m\cdot t$. We experimented with other GSFs for both base and style guidance term, but found that for variably applying different levels of style these very simple GSFs tend to work.

In Fig.~\ref{fig:timevary_types}, ``late start'' affects the layout the least and introduces relatively subtle high frequency style considerations like smooth oil paint texture. ``Base start'' allows for changes to the low frequency components and causes style changes to occur in the pose and the sky, where stars begin to emerge. We can get even more layout from the style prompt component through ``early start''; while this results in a bigger change in the layout of the image, e.g. the cat's pose, it also adds several classical van Gogh elements, e.g. emphasizing the stars in the background.

Additionally, our ``base start'' GSF has very different behavior than constant guidance when the magnitude of the scale is increased. As shown in Figure \ref{fig:timevary_walks}, applying guidance with a constant scale can lead to large image changes as the scale increases. With our default setting for the GSF, the style guidance term is applied primarily to the high-frequency components, weakening the ability for the guidance term to impact the layout or composition but preserving its ability to change texture and middle-frequency components necessary for stylization. This makes the ``walk'' induced by changing the magnitude of the guidance scale smoother perceptually.

\subsubsection{Painting on style with spatial masks}

\begin{figure}[H]
\includegraphics[width=\linewidth]{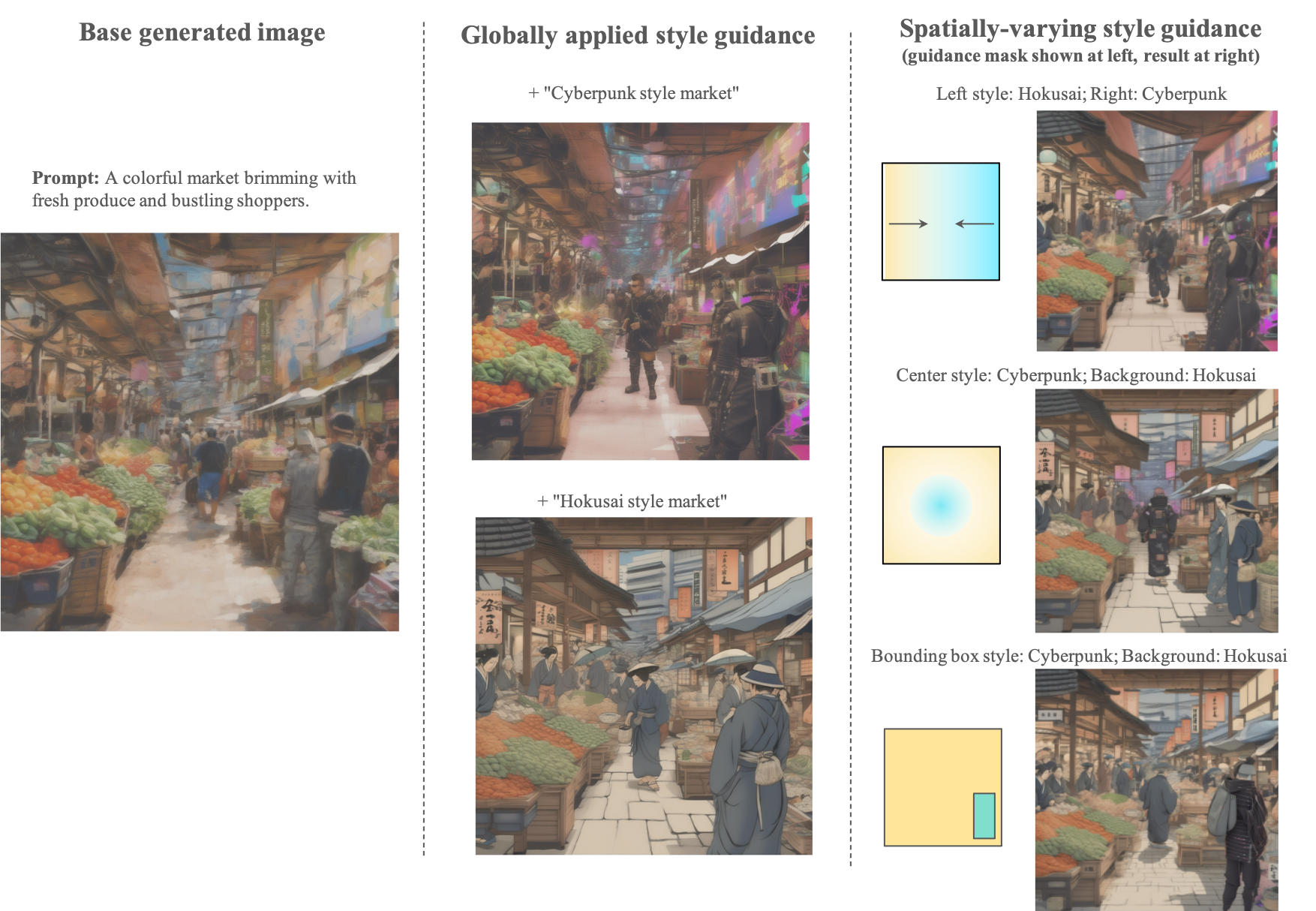}
\caption{\textbf{Spatially-varying guidance functions.} Based on the Base image, style guidance can be applied to different parts of the image. This allows users to define their own masks manually or with computed signals like bounding boxes. Generated with SDXL.}
\label{fig:spacevary}
\end{figure}

Just as we can modulate the guidance scale functions by time, we can also modulate them in pixel space, by writing $s_i = s_i(t,u,v)$. Examples are shown in Figures~\ref{fig:teaser}~and~\ref{fig:spacevary}. The technique flexibly allows us to generate various effects like fading left-to-right from Hokusai to Cyberpunk; doing the same fade radially; or using a bounding box of a semantic object, the person at the bottom right, and changing just this person into cyberpunk.

\section{Experiments}
\begin{figure*}[htb!]
    \hspace{-10pt}\includegraphics[width=0.8\linewidth]{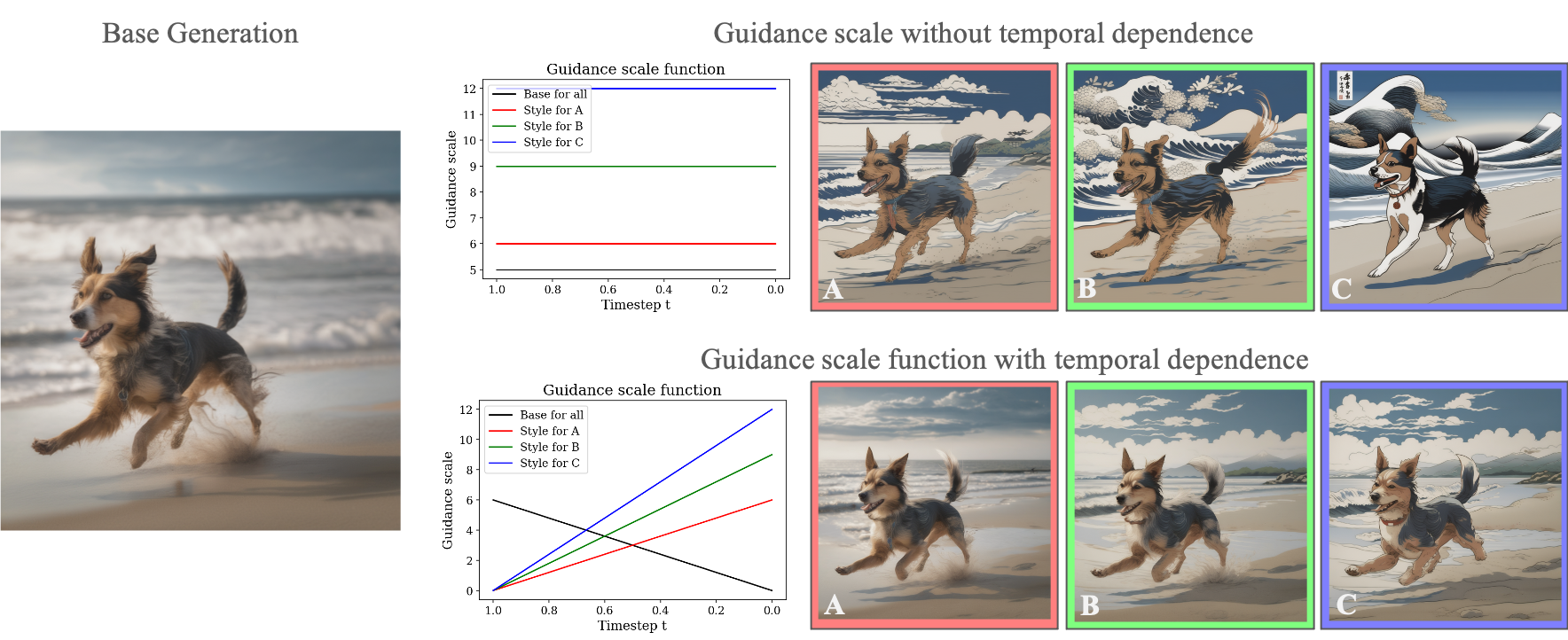}
    \caption{\textbf{Changing the intensity of a style with different guidance scale functions.} We can also select the intensity of the applied style. Changing the magnitude of the guidance scale has different effects based on the temporal component of the GSFs. Increasing the magnitude of a guidance scale without time dependence (bottom row) can change the layout of the image. Our default setting for temporal dependence preserves the rough layout of the base generation while still increasing the amount of  style being added (top row). This is an additional degree of freedom that the GSFs' provide. Base prompt: ``A dog running on a beach''. Added Style prompt: ``Hokusai style dog''. Generated with SDXL.}
    \label{fig:timevary_walks}
\end{figure*}

\subsection{Experimental settings}

For our experiments, we use both Stable Diffusion 1.5 (SD1.5) and Stable Diffusion XL (SDXL), with the default DPM+ and 50 denoising steps. For LORA learning, we conduct model training using 300 screenshots capturing diverse scenes from "Breath of the Wild," employing an instance prompt of "a photo of [$\mathcal{V}$] style." The learning rate was set to $10^{-4}$ for 500 steps.

For DB training, we follow the same experimental procedure used the \cite{ruiz2023dreambooth} paper with some minor hyperparameter adjustments made to change from Imagen~\cite{saharia2022photorealistic}.%
\ifNotarXiv{(See Supplemental material)}
Specifically, we train our models on 30 objects from the Dreambooth dataset~\cite{ruiz2023dreambooth} using SD1.5. The training process generates 1,000 samples from "a photo of [class noun]" class prior, and uses a batch size of 1, $\lambda = 1$, a learning rate of $10^{-6}$, and a total of 1,500 iterations.

\subsection{Walks in Style and Personalization}

In Figures \ref{fig:singlestylemanyapplications1} and \ref{fig:singlestylemanyapplications2}, we show straight forward applications of style to a given layout. These examples use different architectures (SD1.5 and SDXL) and a diverse set of styles including a LORA-trained style for the popular video game ``Breath of the Wild''. We also demonstrate in Figure \figmixing~  that our technique is not limited to one or even two styles and can be used with an arbitrary number of guidance terms, mixing 4 distinct styles globally in the same image, taking some aspects from each.

Figure \figstylegridmacaron{} shows that we can control the amount of style applied to a base image while preserving the rough layout, even when using a style trained with LORA on SD1.5. This example applies the style guidance term on the full image but uses the ``default'' option for the temporal aspect of the guidance scale function. The top left image is the base generation without any stylization applied; lower in the grid has more ``Breath of the Wild'' style applied to it, and to the right has more ``Qi-Bashi calligraphy style''. Figure \ref{fig:dbteaser} shows the same capability for personalization on SD1.5, amplifying either the subject (the specific type of dog) or the action and setting. Note that amplifying both still preserves the subject while forcing a change in pose / action. In Figure \figdbxstyle, we show that both of these terms can be applied at once; we can take a DB subject and incrementally add style. Despite the addition of the style, the guidance term for personalization preserves the shape and color of the subject.

Our method also allows users to interpolate between multiple styles. In Figure \ref{fig:smoothinterp} and \ref{fig:interpxl}, we show that our smooth interpolation between two different styles for models with very different architectures SD1.5 and SDXL. We also complete a user study in the following section on interpolation.

\subsection{Baselines}

While stylization is a popular topic in Computer Vision, much of the prior work focused on walking in the latent space of GANs. 

In the process of interpolating styles, we benchmark against two widely adopted techniques in the community, employing the compel library \cite{Compel:GitHub}: prompt blending and prompt weighting. To achieve blending, such as between Picasso and Hokusai, we assign weights $\lambda$ and $1-\lambda$ to the prompts "Artist painting vivid sunset on beach canvas in the style of Picasso" and "Artist painting vivid sunset on beach canvas in the style of Hokusai," respectively, with $\lambda \in [0,1]$. Regarding prompt weighting, we consolidate the two prompts into a singular prompt, "Artist painting vivid sunset on beach canvas in the style of Picasso and Hokusai," and apply weights $\lambda$ and $1-\lambda$ to the terms "Picasso" and "Hokusai," respectively, where $\lambda \in [0, k]$. We observe that when $k=1$, the change is nearly imperceptible, prompting our choice of $k=1.5$ for a more distinct transition. The outcomes, displayed in Figure~\ref{fig:smoothinterp}, reveal that both baseline methods yield more abrupt transitions. Notably, in certain instances, prompt blending leads to the complete loss of style, a phenomenon we term the 'style-free valley,' which is not observed with \ourname{}.

\subsection{User study}

Since there is no clear source of ground truth for our task, we conducted a user study to determine if the results we generate are viewed as aesthetically pleasing. 
Users were presented with a start image and an end image, along with two possible series of images showing the transition from the start to the end, similar to Figure~\ref{fig:smoothinterp}.
One series was generated by the Compel baseline and one by DreamWalk; we randomly picked the layout in which the two series were displayed.
Users were asked which series produced a ``more pleasingly smooth transition''. Our user study closely followed \cite{ruiz2023dreambooth},
\ifNotarXiv{(Details are provided in the supplemental material.)} 
Users preferred the DreamWalk output 68\% of the time.

\section{Limitations and Ethics}
One of the main limitations of this technique is global entanglements as discussed in Section \ref{sec:deconstructingprompts}. While we show that this can be mitigated through prompt engineering and guidance scale functions, not all the correlations can be prevented. Additionally, at inference time our method requires one forward pass per prompt component, unlike fine-tuning methods (although of course the latter require the actual fine-tuning time). We also acknowledge that this work, along with most generative media, comes with a degree of ethical concern. Better control over generated images can increase the risk of these tools being used by bad actors.

\section{Conclusion}
We have presented \ourname, a generalized guidance formulation designed specifically for personalizing text-to-images. This method allows for granular control over the amount of style applied or adherence to a DB token or LORA. We have empirically shown the efficiency of this method on several tasks including style interpolation, DB sampling, changing materials, and granularly manipulating the texture and layout of generated images.

\ifarXiv{
\subsection{Acknowledgments}
The portion of this work performed at Cornell has been supported by a gift from the Simons Foundation. We thank Yuanzhen Li for coming up with the project name, and Miki Rubinstein and Deqing Sun for helpful discussions.
}
\begin{figure*}[t!]
    \includegraphics[width=0.95\textwidth]{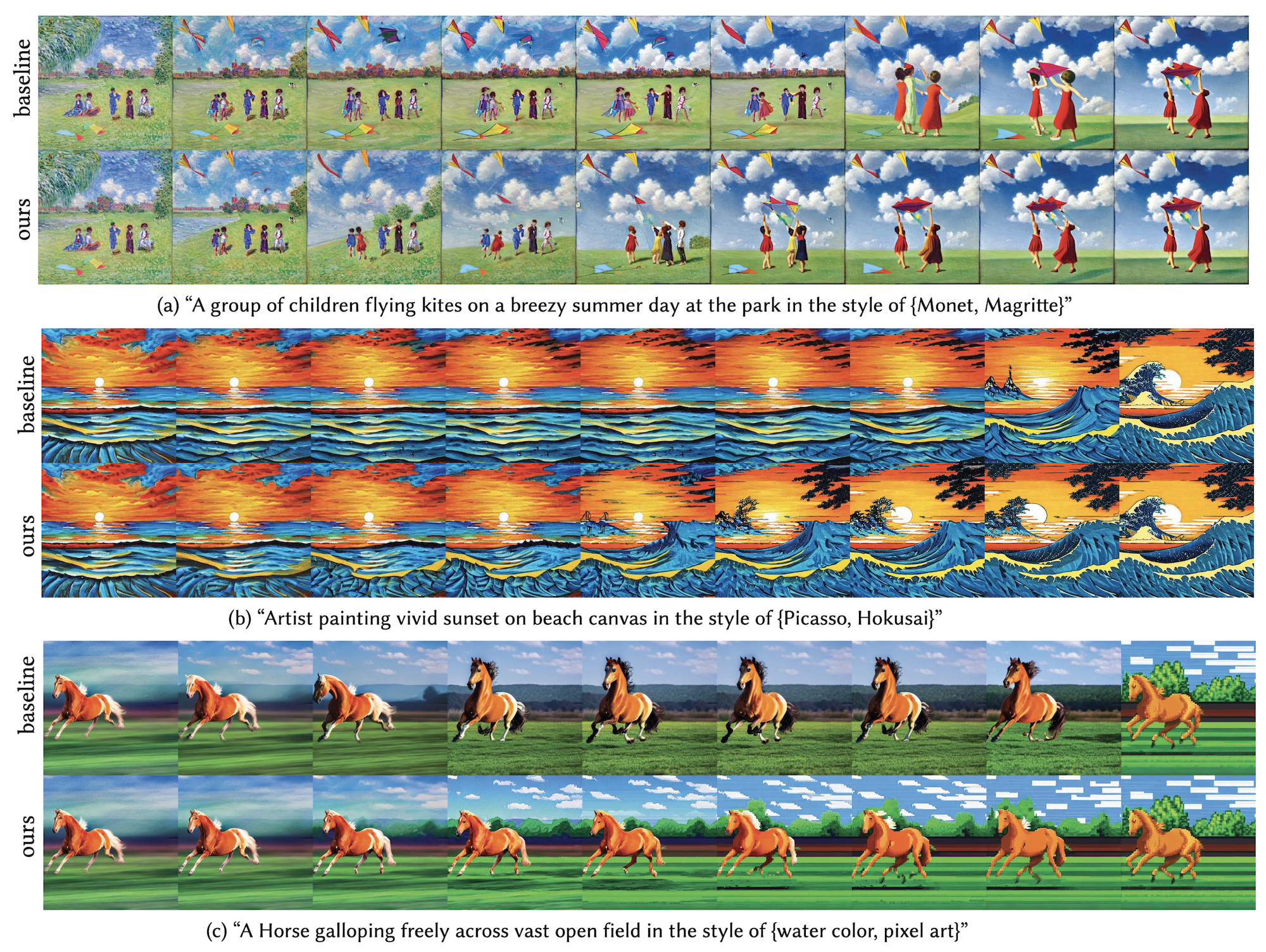}
    \ignore{
    \newcommand{\sidelabel}{\rotatebox{90}{\hspace{0.5cm} ours\hspace{1cm} baseline}}
    \centering
    \begin{subfigure}[c]{0.9\textwidth}
    \centering
    \begin{tabular}{cc}
    \centering
    \setlength\tabcolsep{1.5pt}
    \sidelabel &\includegraphics[width=\textwidth]{images_in_paper/images/smoothlyvarying/24305-children.png}
    \end{tabular}\vspace{-6pt}
    \caption{``A group of children flying kites on a breezy summer day at the park in the style of \{Monet, Magritte\}''}
    \end{subfigure} \\
    \vspace{4pt}
    
    \begin{subfigure}[c]{0.9\textwidth}
    \centering
    \begin{tabular}{cc}
    \centering
    \setlength\tabcolsep{1.5pt}
    \sidelabel & \includegraphics[width=\textwidth]{images_in_paper/images/smoothlyvarying/88200-sunset.png}
    \end{tabular}\vspace{-6pt}
    \caption{``Artist painting vivid sunset on beach canvas in the style of \{Picasso, Hokusai\}''}
    \end{subfigure} \\
    \vspace{4pt}
    
    \begin{subfigure}[c]{0.9\textwidth}
    \centering
    \begin{tabular}{cc}
    \centering
    \setlength\tabcolsep{1.5pt}
    \sidelabel & \includegraphics[width=\textwidth]{images_in_paper/images/smoothlyvarying/69240-horse.png}
    \end{tabular}\vspace{-6pt}
    \caption{``A Horse galloping freely across vast open field in the style of \{water color, pixel art\}''}
    \end{subfigure}
    }
    \caption{Baseline comparisons, in each panel prompt blending~\cite{promptweighting} shown at the top and
    \ourname{} shown below. 
    Prompt blending makes an abrupt transition while \ourname{} changes smoothly; prompt blending also sometimes drops both styles altogether. Generated with SD1.5.}
    \label{fig:smoothinterp}
\end{figure*}

\begin{figure}[b]
\newcommand{\macwidth}{0.24\linewidth}
\setlength\tabcolsep{0.5pt}
\begin{tabular}{cccc}
\tiny{Breath of the Wild LORA} & \tiny{Hokusai} & \tiny{Pixel Art} & \tiny{Watercolor}\\
\includegraphics[width=\macwidth]{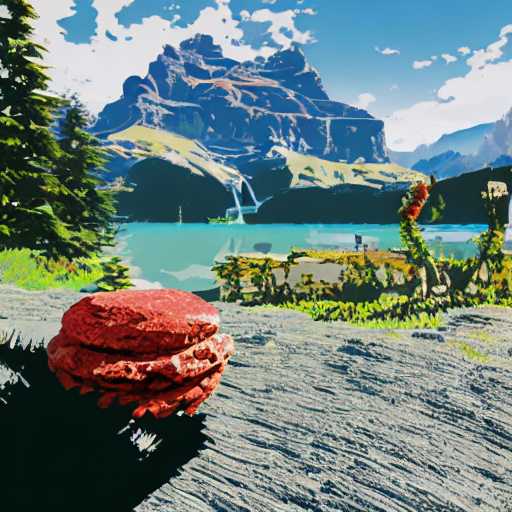} &
\includegraphics[width=\macwidth]{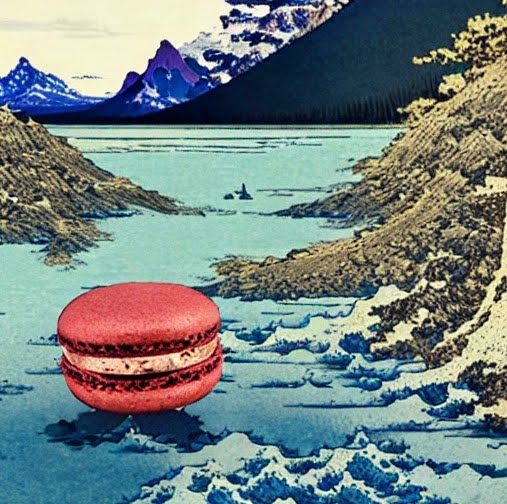} &
\includegraphics[width=\macwidth]{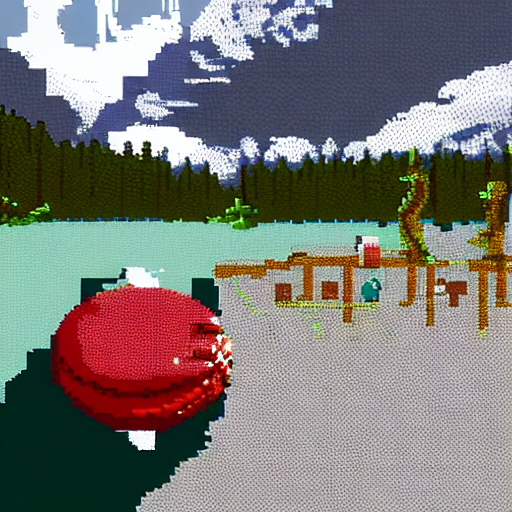} &
\includegraphics[width=\macwidth]{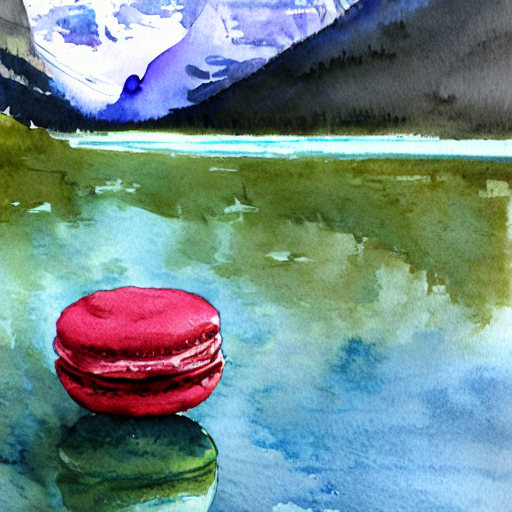} \\
\includegraphics[width=\macwidth]{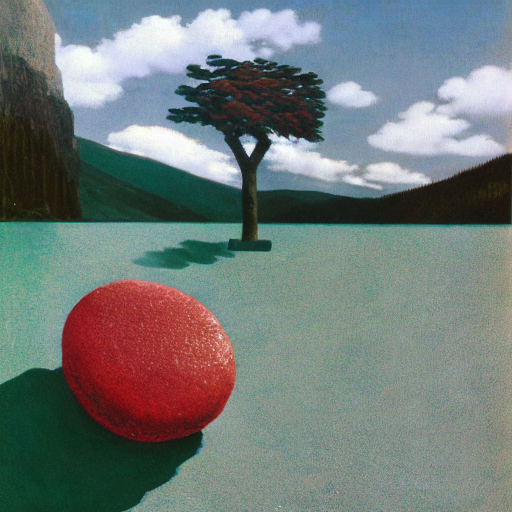} &
\includegraphics[width=\macwidth]{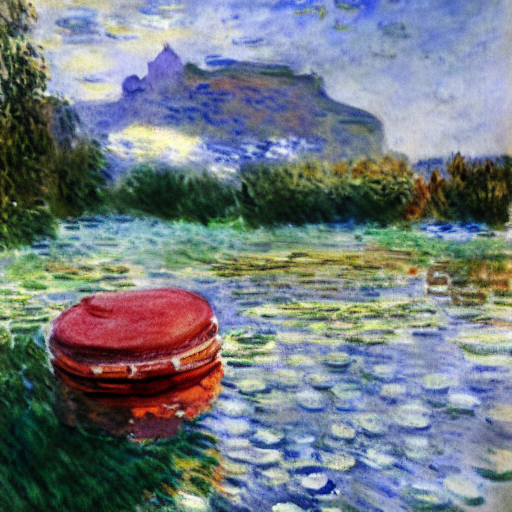} &
\includegraphics[width=\macwidth]{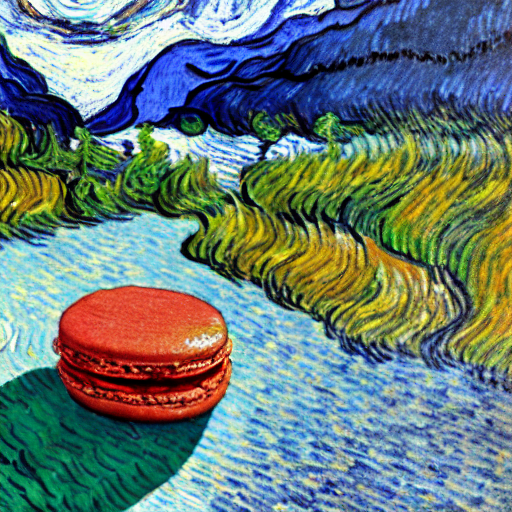} &
\includegraphics[width=\macwidth]{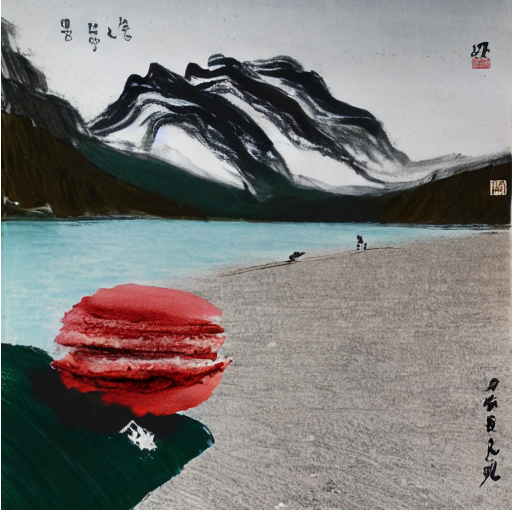} \\
\tiny{Magritte} & \tiny{Monet} & \tiny{Van Gogh} & \tiny{Qi Bashi}
\end{tabular}
    \caption{Our style application works across a wide variety of styles, including those already defined in the model's text and those trained using LORA. {Prompt: ``A macaron in Banff National Park''}. Generated with SD1.5.}\label{fig:singlestylemanyapplications1}
\end{figure}

\begin{figure}[b]
\newcommand{\macwidth}{0.24\linewidth}
\setlength\tabcolsep{0.5pt}
\begin{tabular}{cccc}
\tiny{Pixel Art} & \tiny{Monet} & \tiny{Qi Bashi} & \tiny{Hokusai}\\
\includegraphics[width=\macwidth]{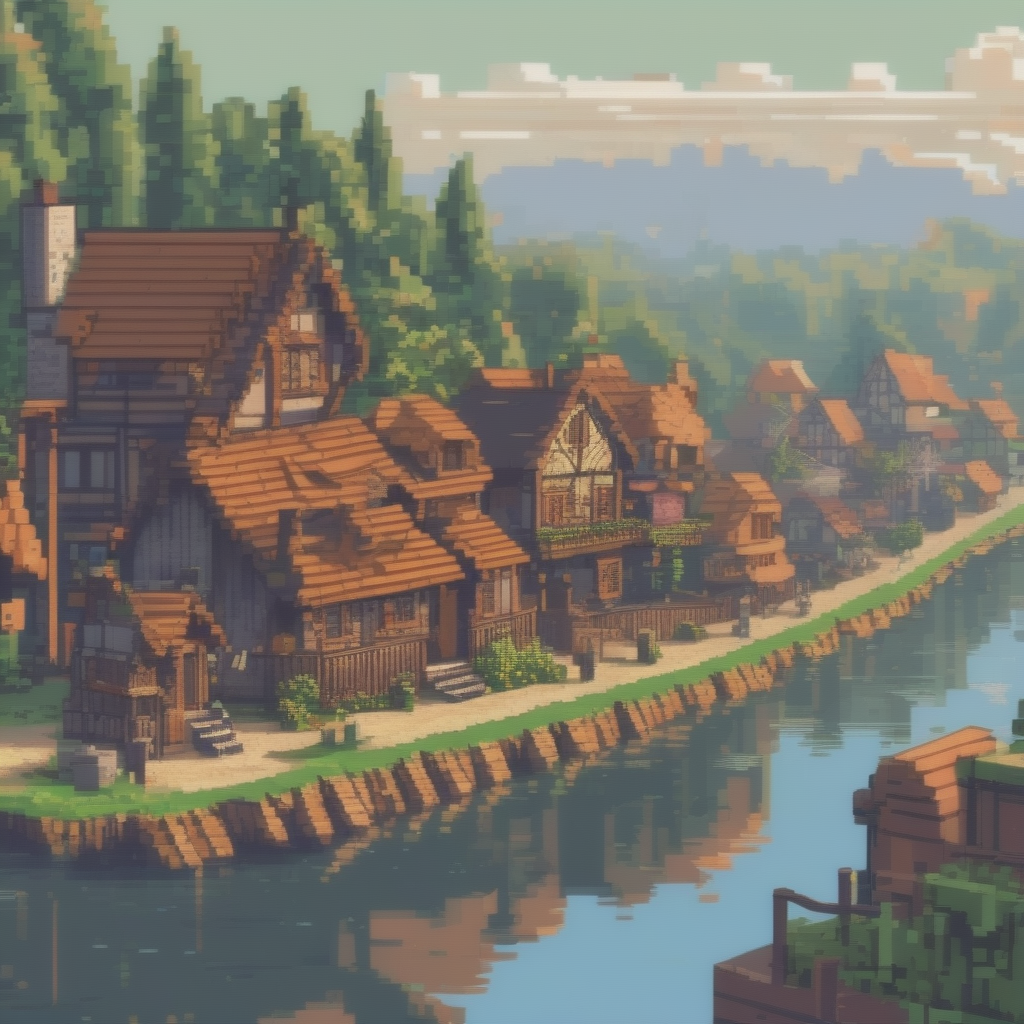} &
\includegraphics[width=\macwidth]{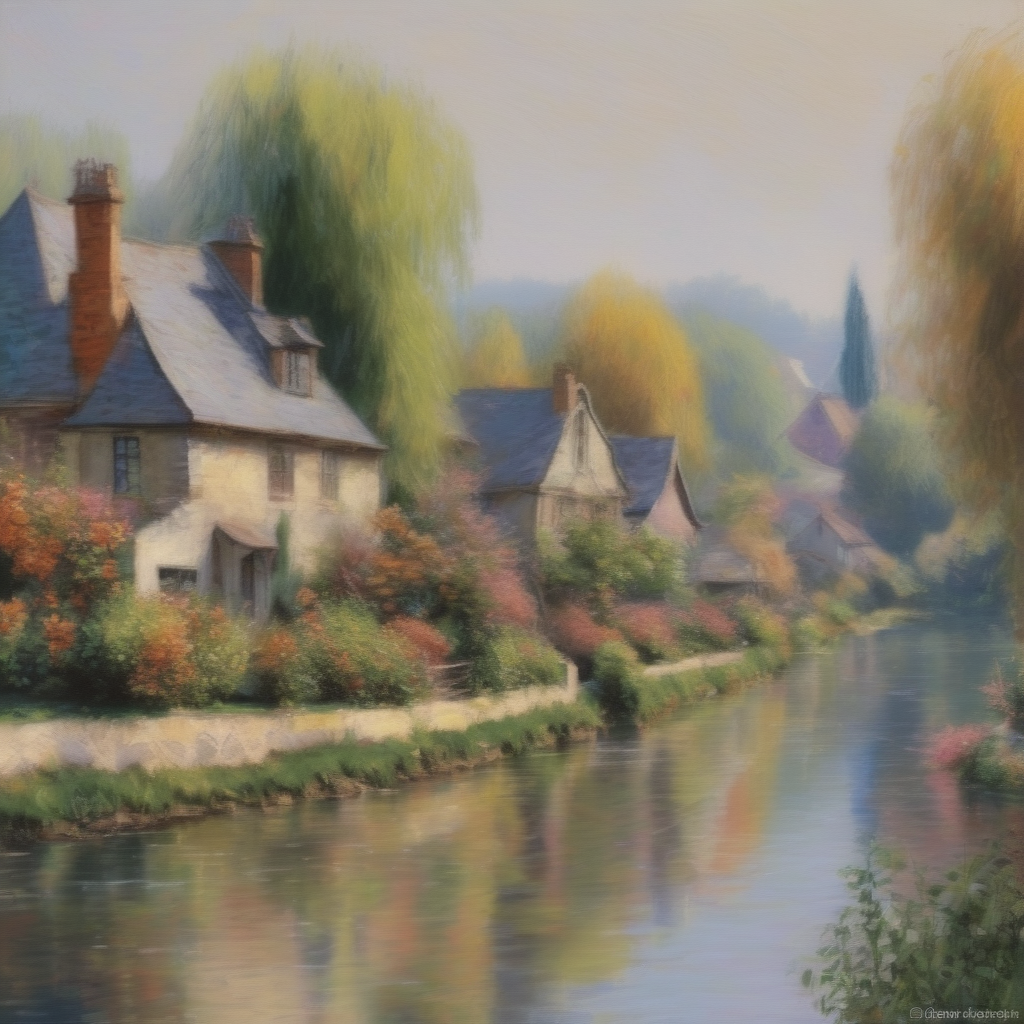} &
\includegraphics[width=\macwidth]{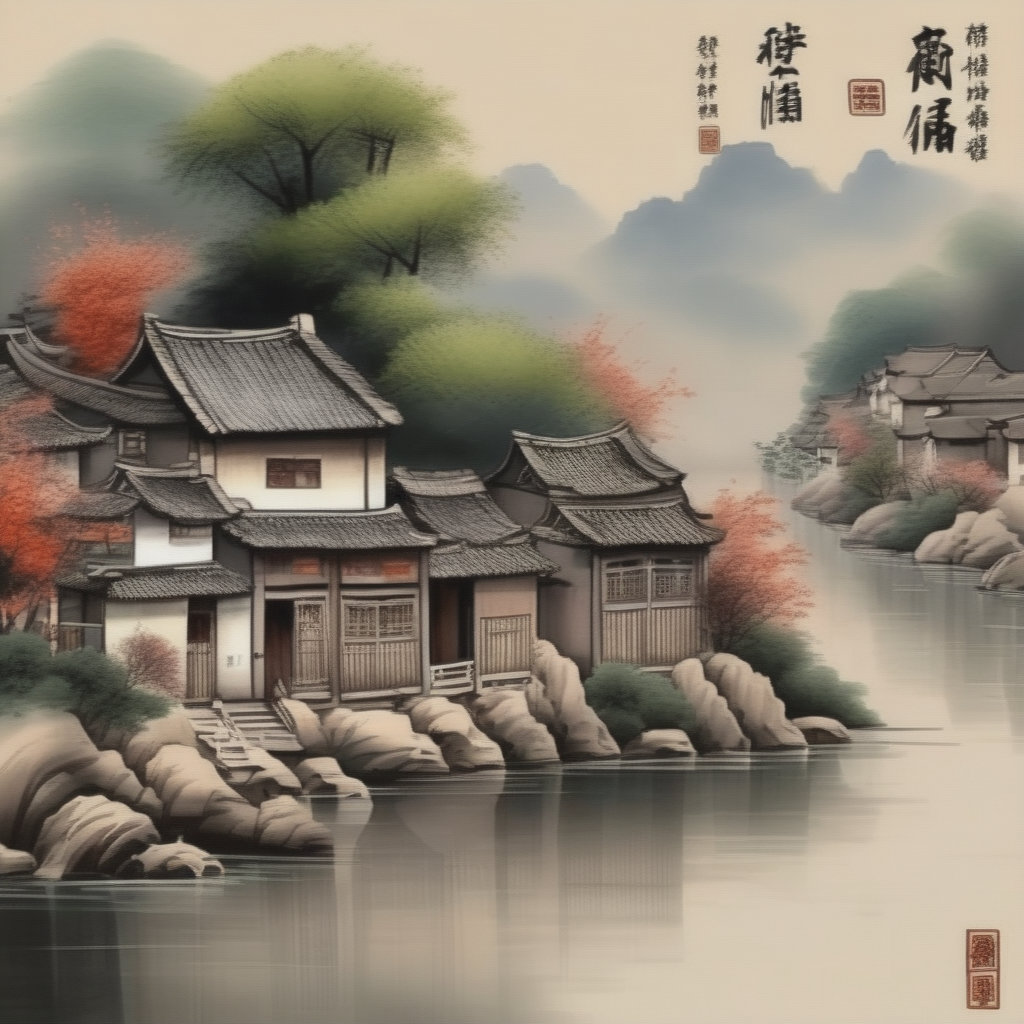} &
\includegraphics[width=\macwidth]{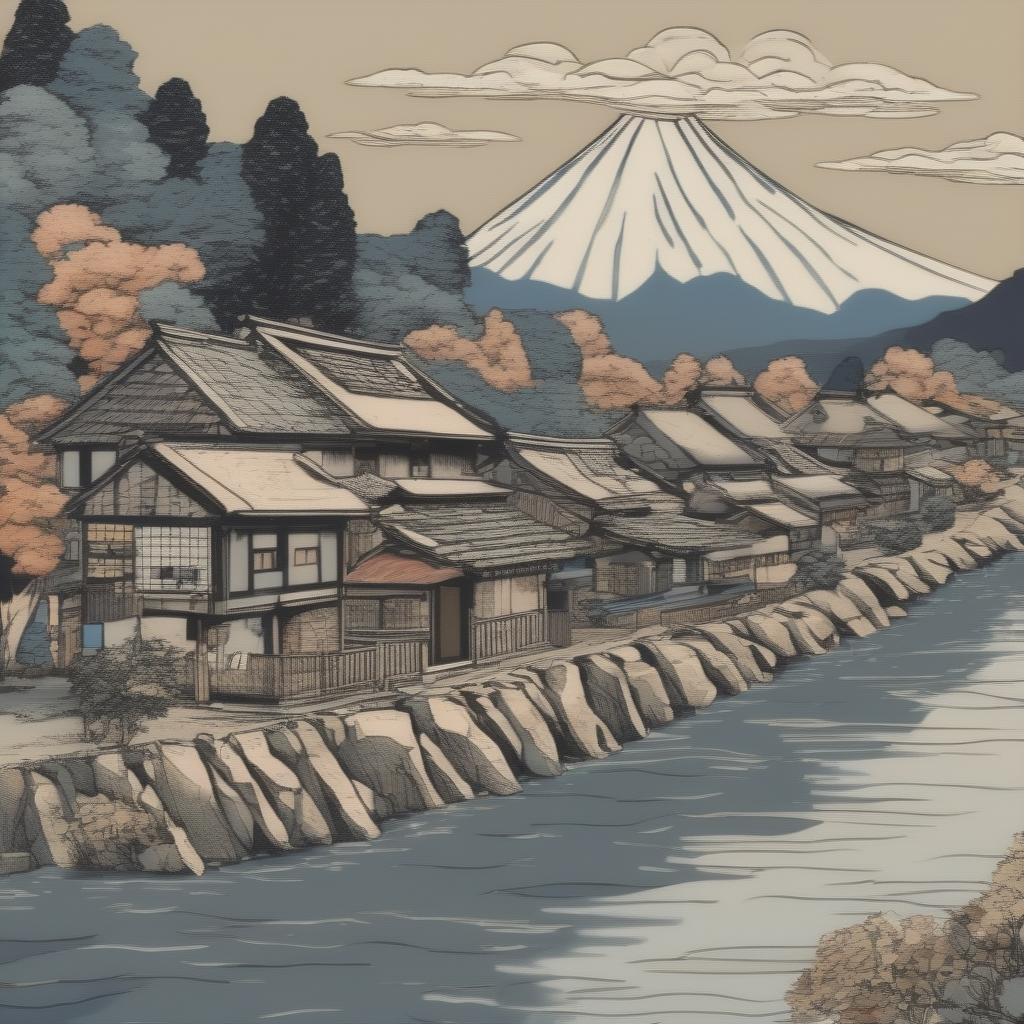} \\
\includegraphics[width=\macwidth]{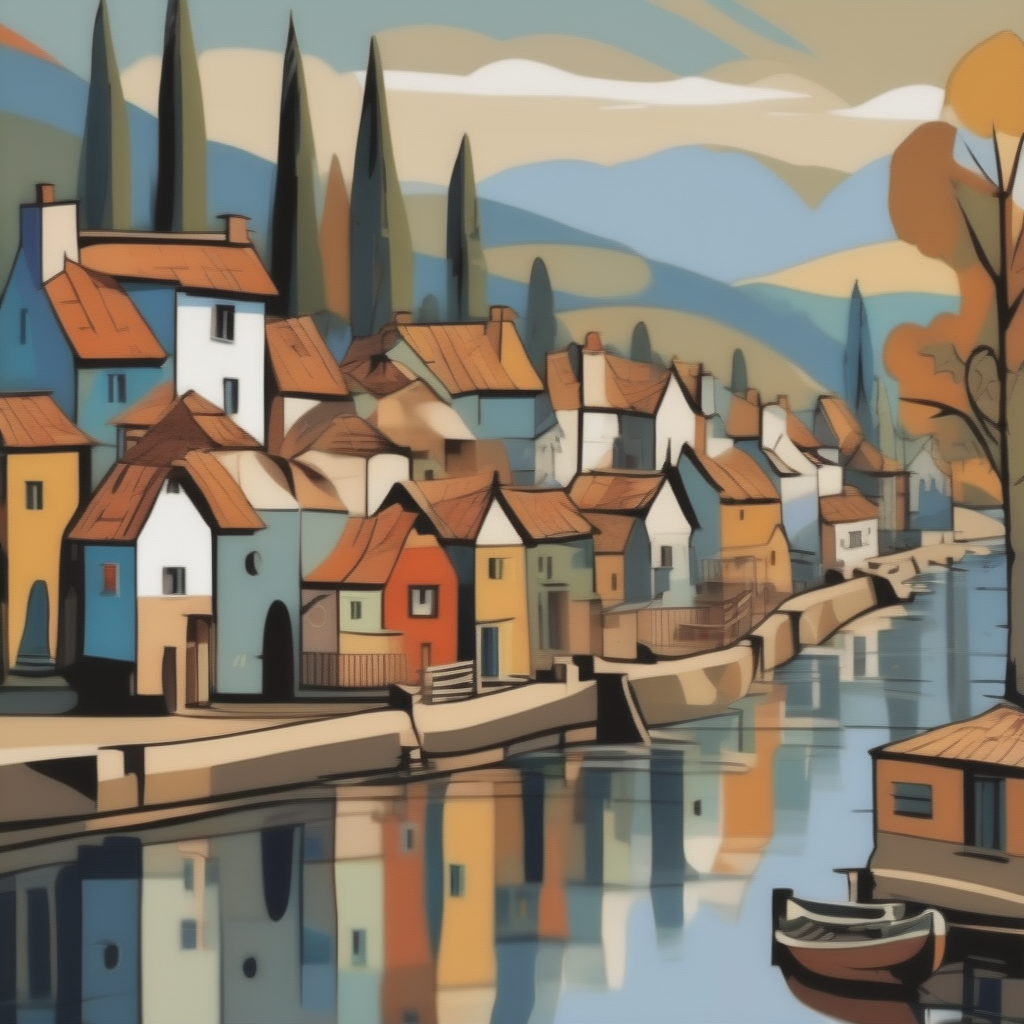} &
\includegraphics[width=\macwidth]{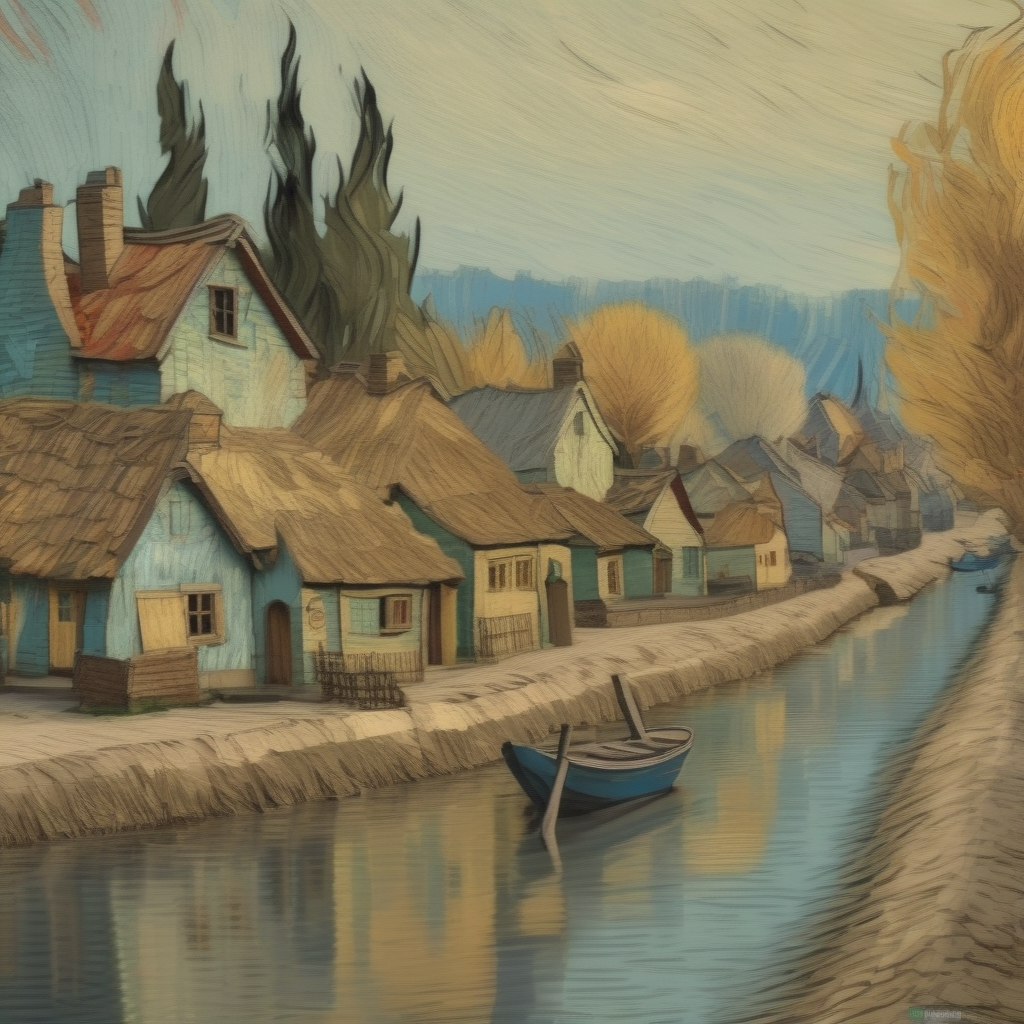} &
\includegraphics[width=\macwidth]{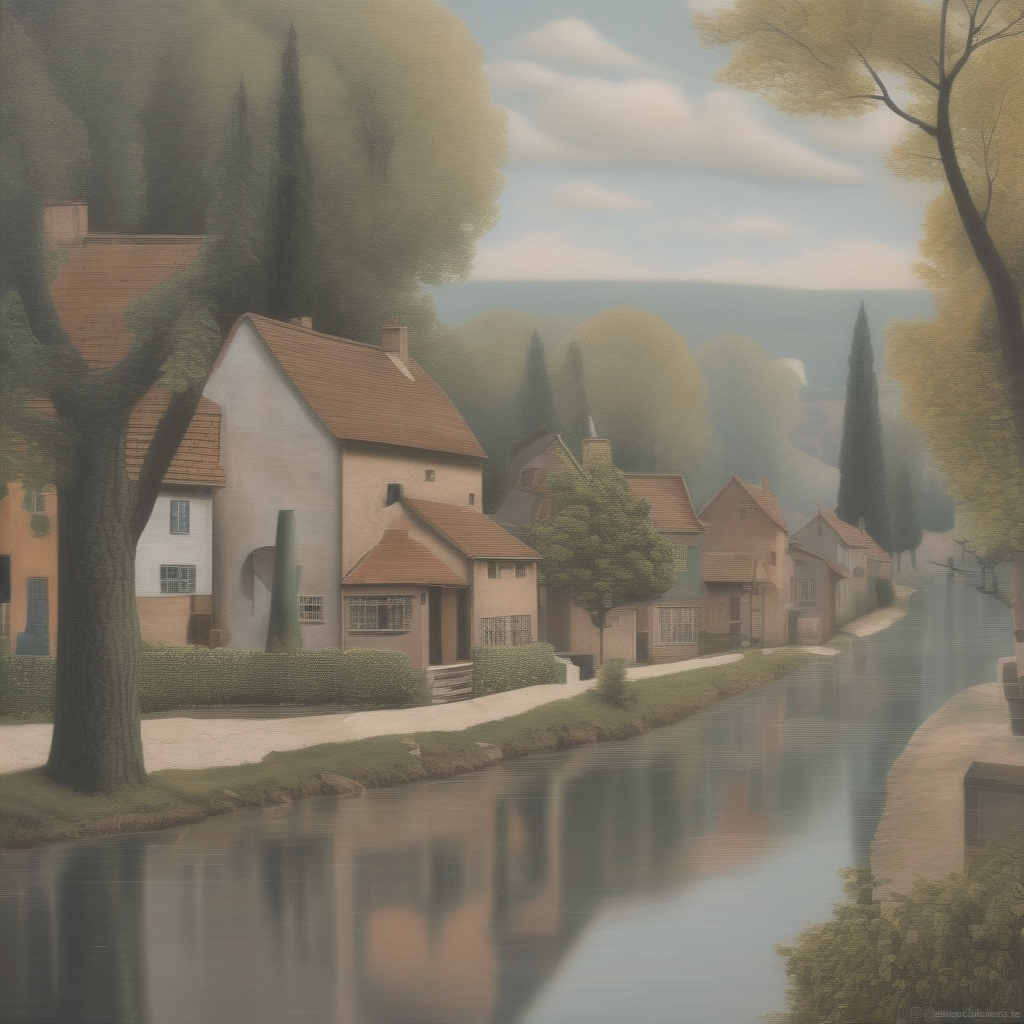} &
\includegraphics[width=\macwidth]{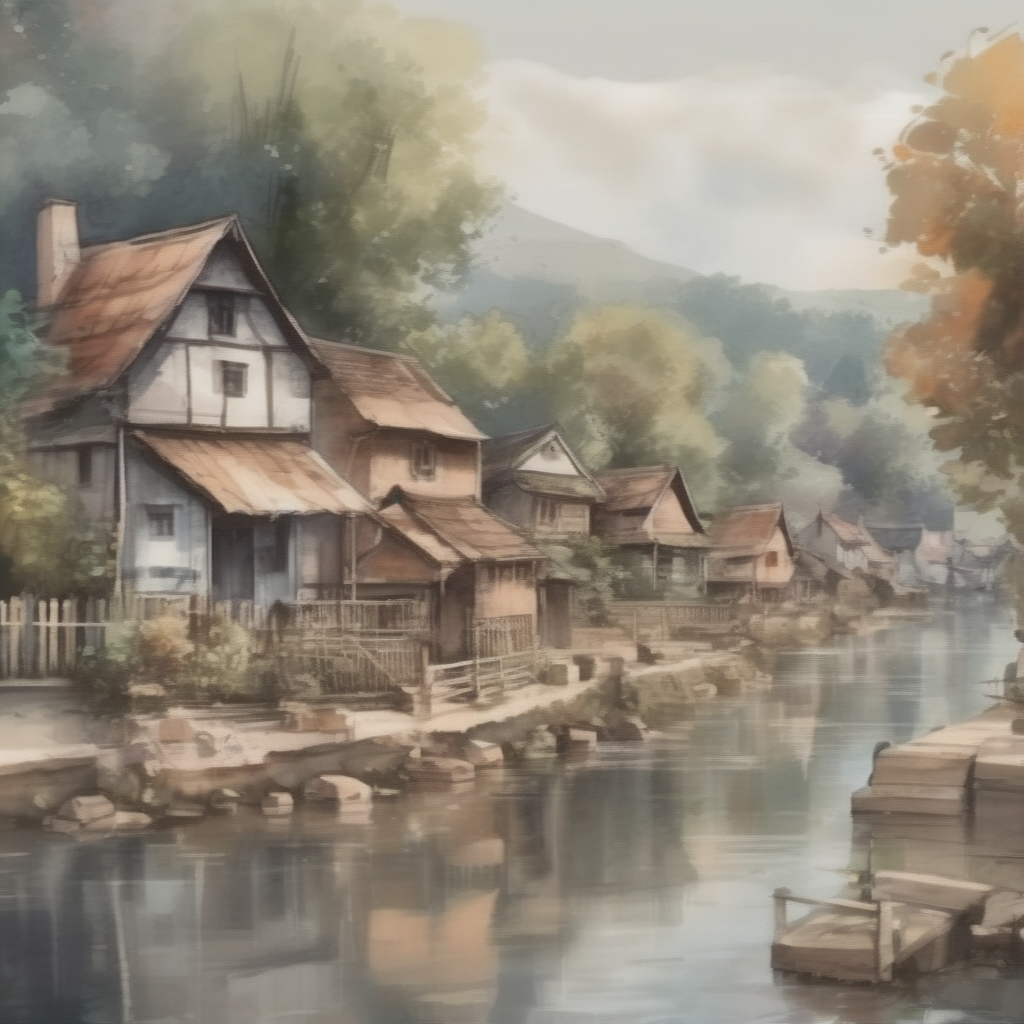} \\
\tiny{Picasso} & \tiny{van Gogh} & \tiny{Magritte} & \tiny{Watercolor}
\end{tabular}
    \caption{Our style application works on any diffusion model regardless of architecture differences.
    {Prompt: ``A peaceful riverside village with charming old cottages''}. Generated with SDXL.}\label{fig:singlestylemanyapplications2}
\end{figure}

\begin{figure*}
\setlength{\tabcolsep}{0.4pt}
\begin{tabular}{ccccccc}

\raisebox{35pt}[0pt][0pt]{\rotatebox{90}{\footnotesize Van Gogh}} &
\includegraphics[width=0.19\textwidth]{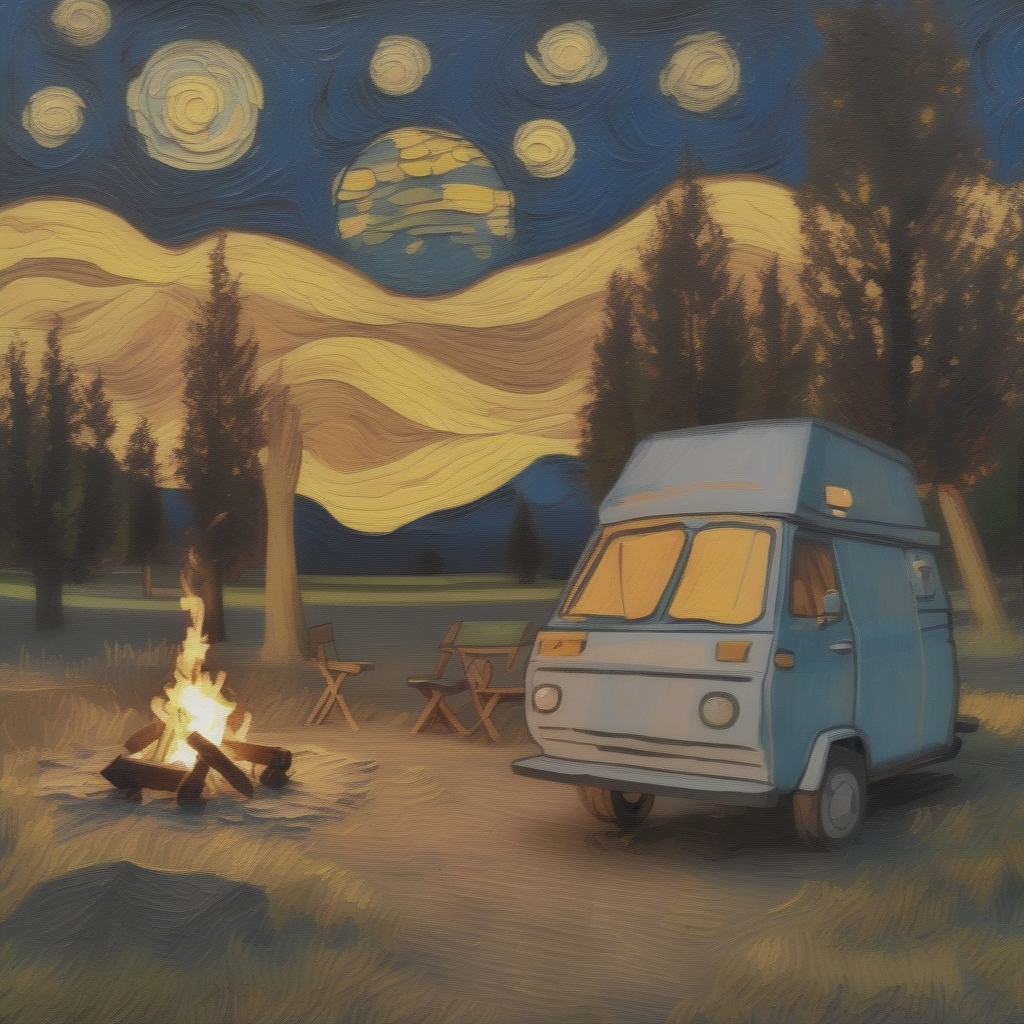} &
\includegraphics[width=0.19\textwidth]{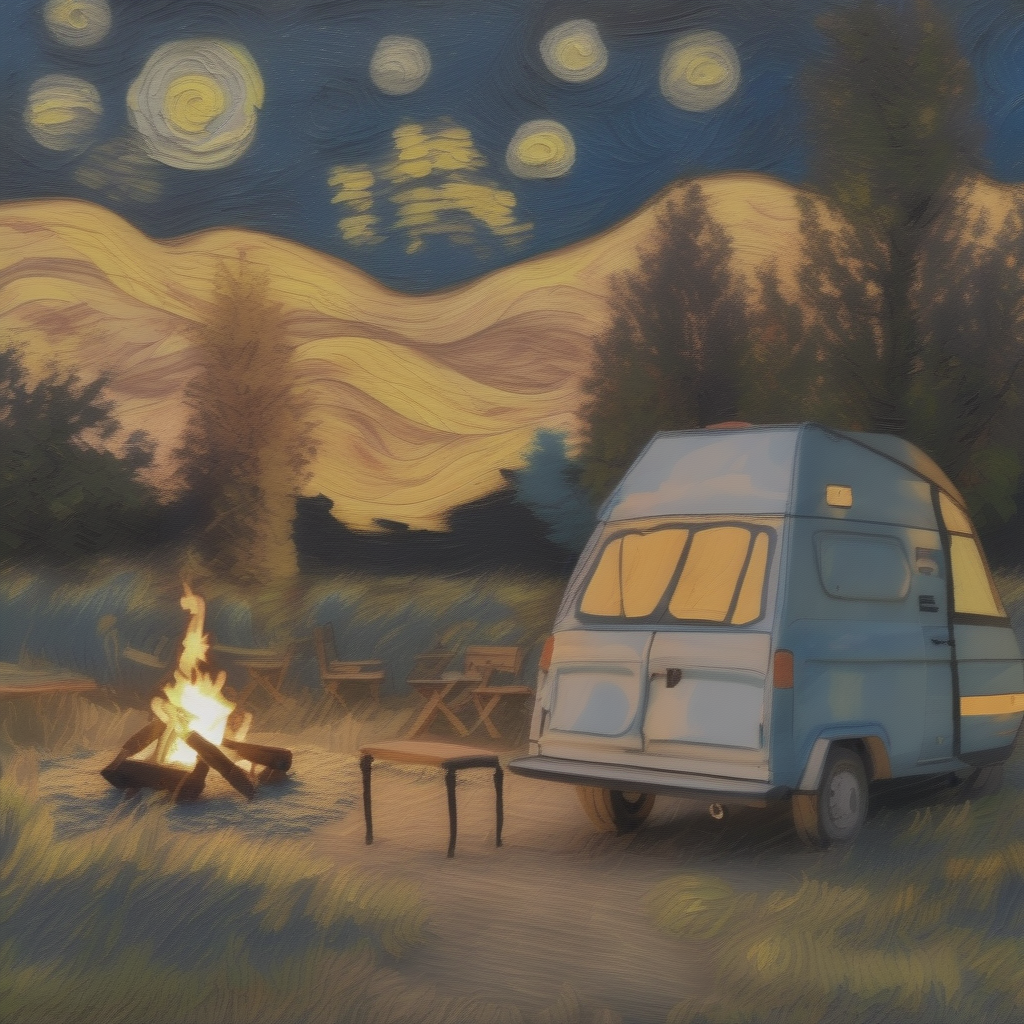} &
\includegraphics[width=0.19\textwidth]{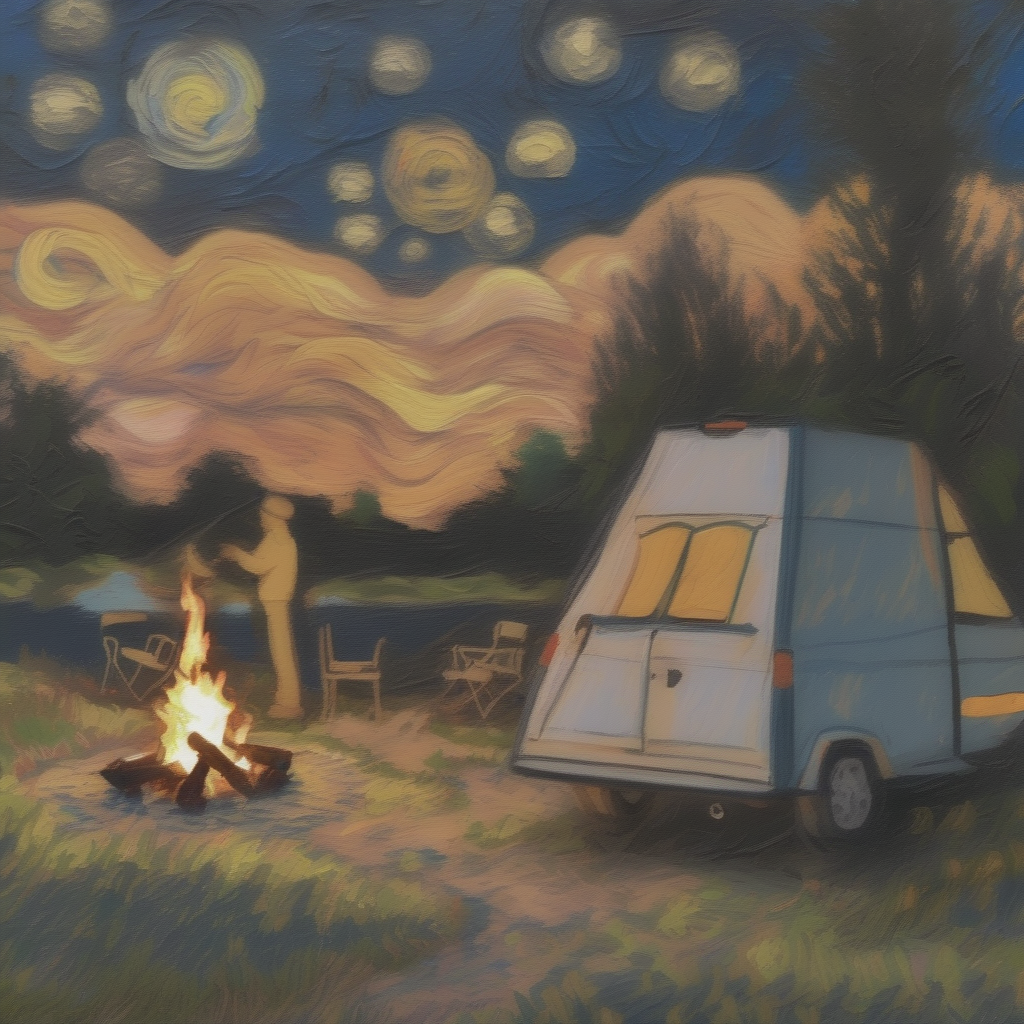} &
\includegraphics[width=0.19\textwidth]{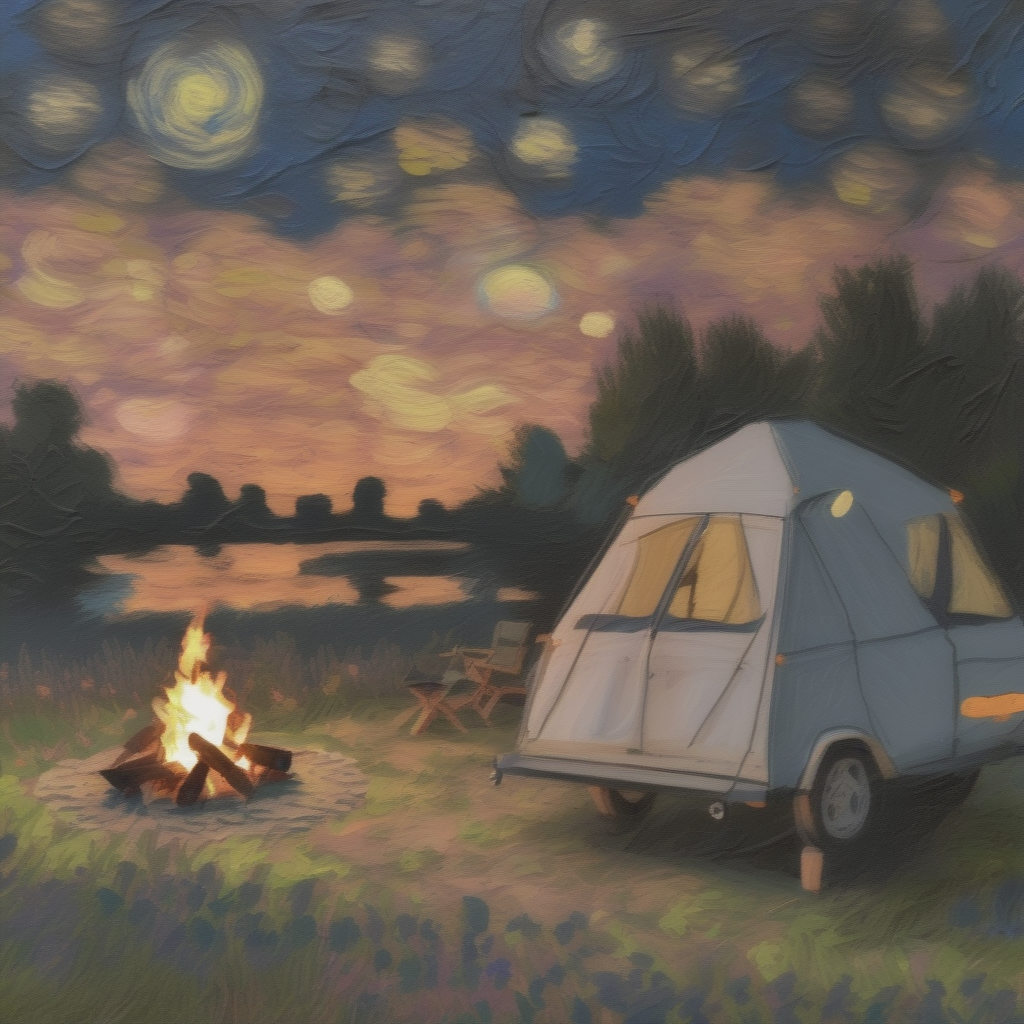} &
\includegraphics[width=0.19\textwidth]{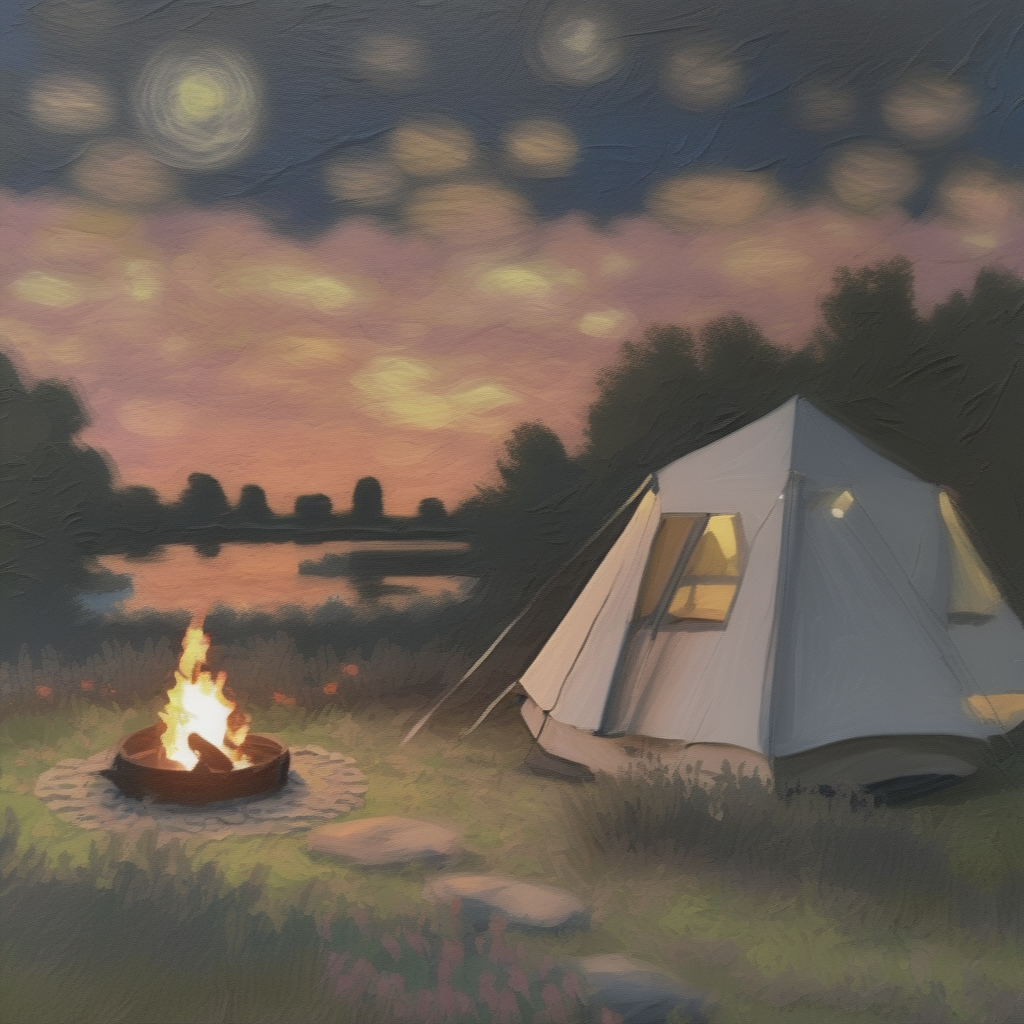} & 
\raisebox{60pt}[0pt][0pt]{\rotatebox{270}{\footnotesize  Monet}} \\


\raisebox{35pt}[0pt][0pt]{\rotatebox{90}{\footnotesize Van Gogh}} &
\includegraphics[width=0.19\textwidth]{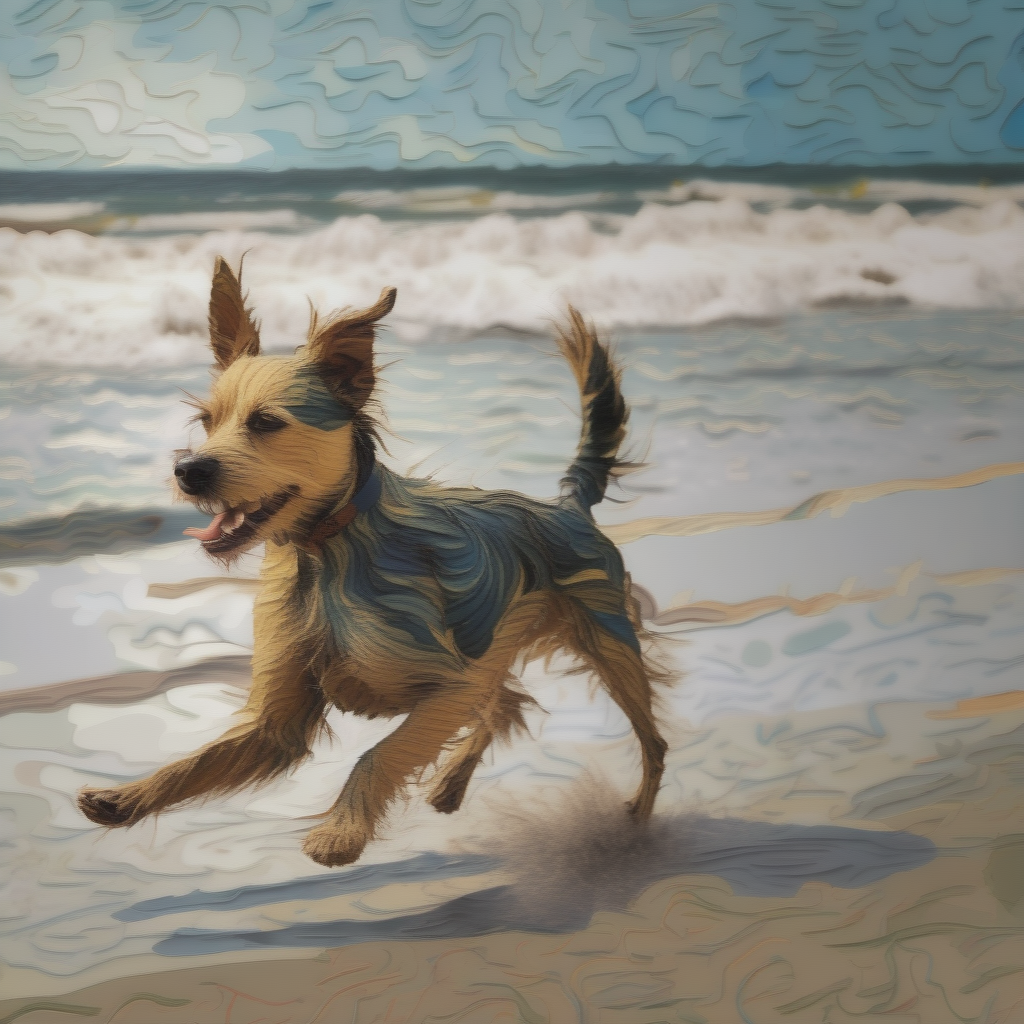} &
\includegraphics[width=0.19\textwidth]{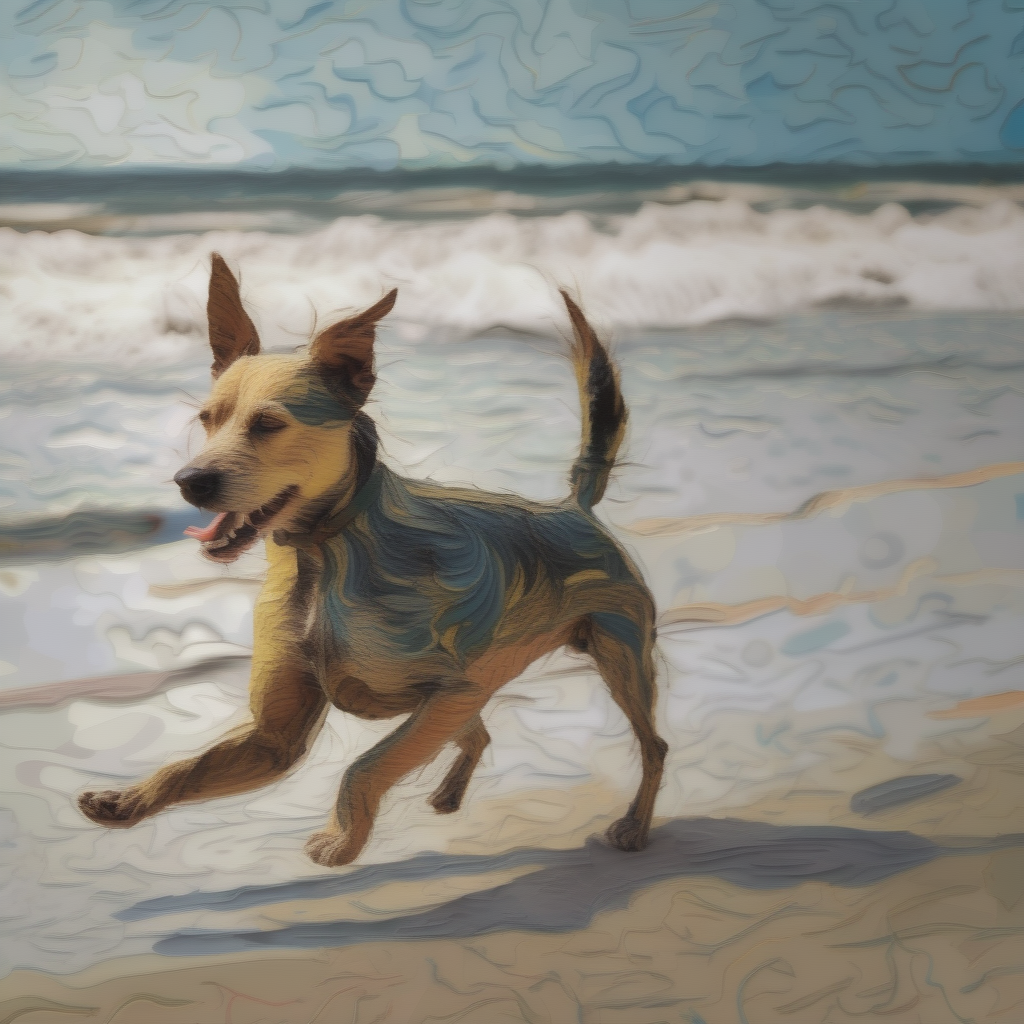} &
\includegraphics[width=0.19\textwidth]{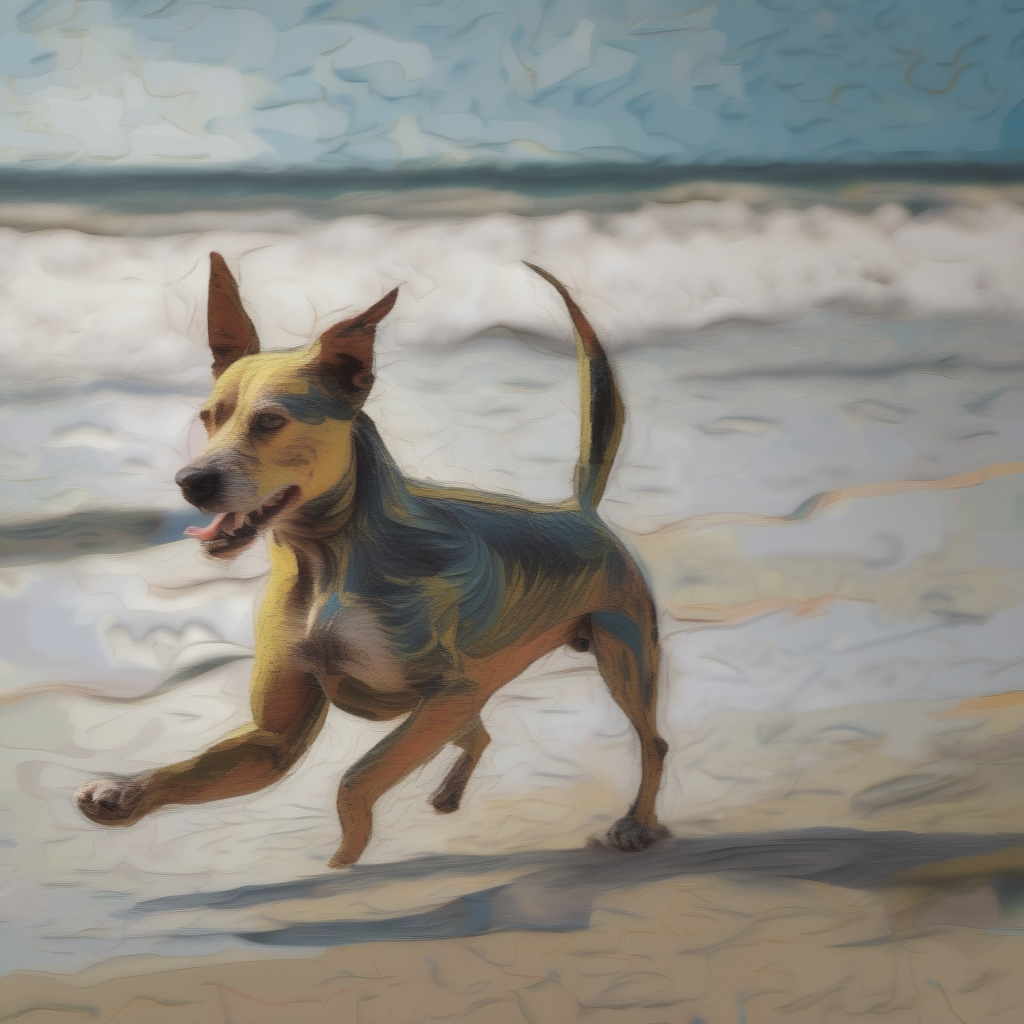} &
\includegraphics[width=0.19\textwidth]{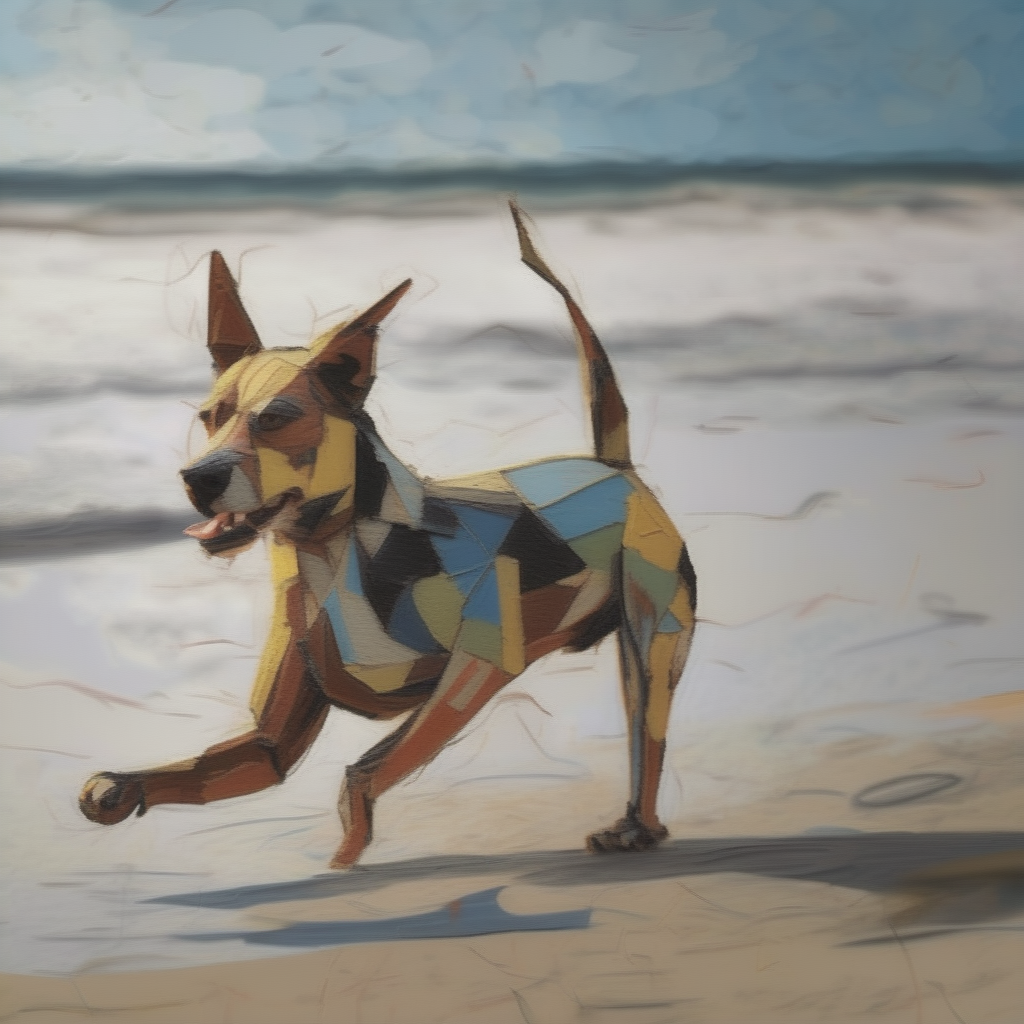} &
\includegraphics[width=0.19\textwidth]{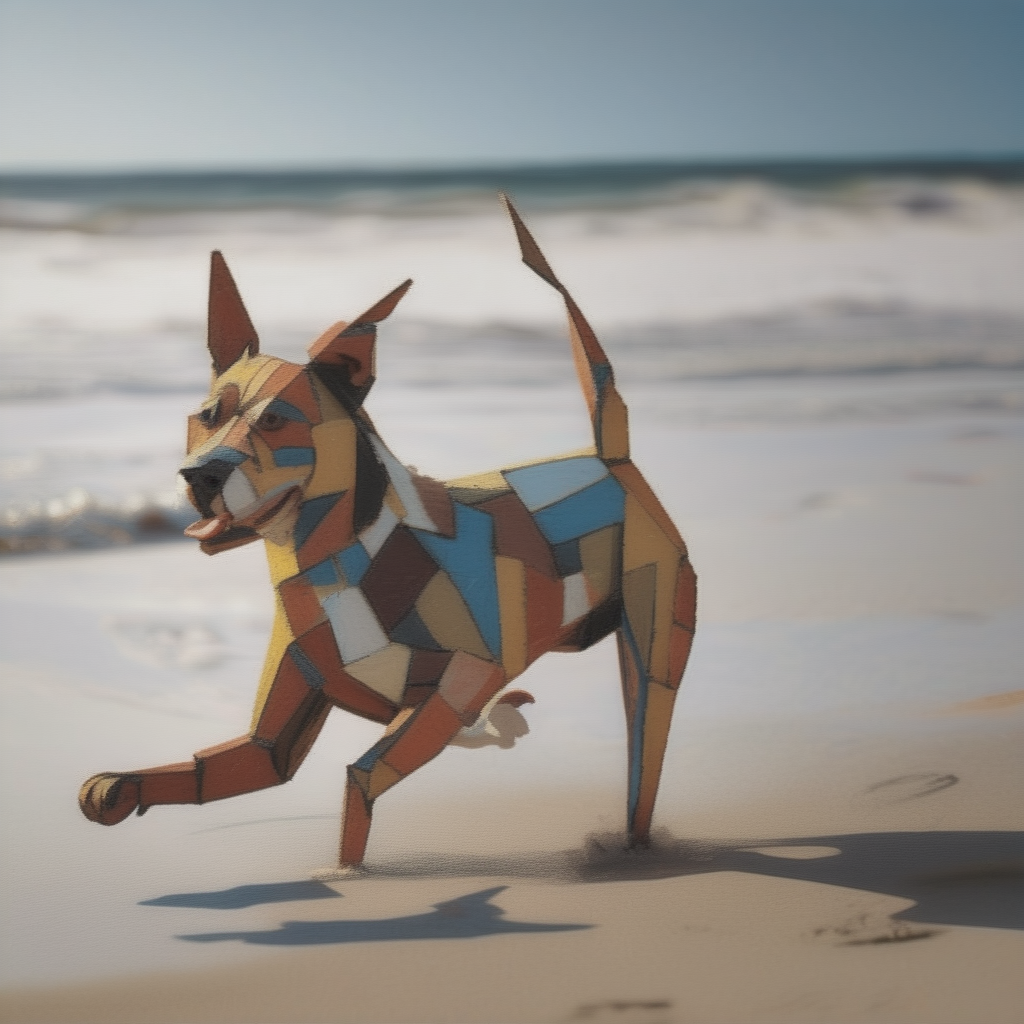} & 
\raisebox{60pt}[0pt][0pt]{\rotatebox{270}{\footnotesize Picasso}} \\

\end{tabular}
\caption{Our method can interpolate between two different styles while maintaining the rough layout of the scene. Top row is ``a campsite with a fire at night''. Bottom row is ``a dog running on a beach''. Generated with SDXL.}
\label{fig:interpxl}
\end{figure*}

\hspace{-10pt}\begin{figure*}
\includegraphics[width=\linewidth]{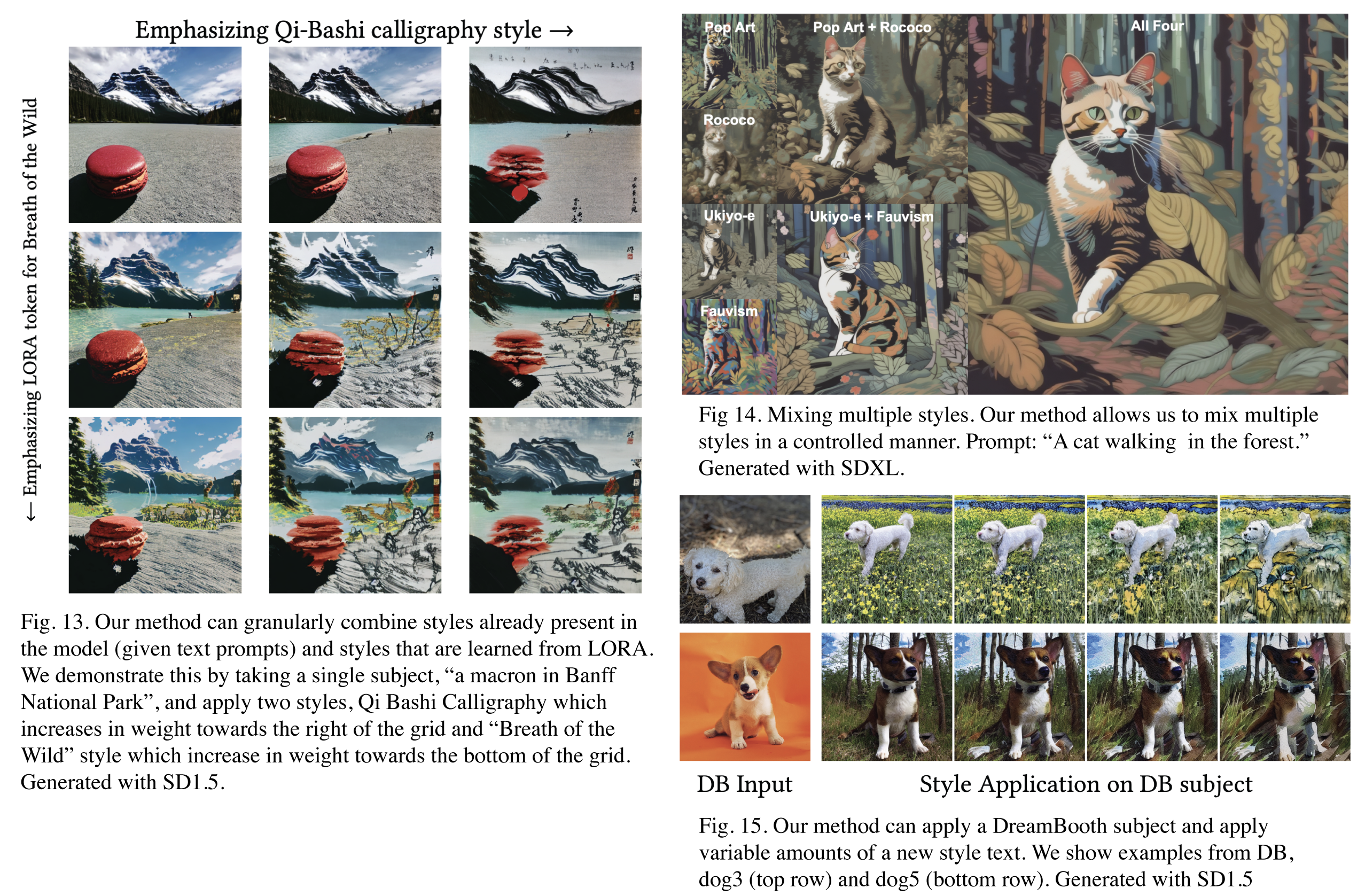} \label{fig:mixing}\end{figure*}

\clearpage

{
\bibliographystyle{ACM-Reference-Format}
\bibliography{reference}


\begin{thebibliography}{51}


\ifx \showCODEN    \undefined \def \showCODEN     #1{\unskip}     \fi
\ifx \showDOI      \undefined \def \showDOI       #1{#1}\fi
\ifx \showISBNx    \undefined \def \showISBNx     #1{\unskip}     \fi
\ifx \showISBNxiii \undefined \def \showISBNxiii  #1{\unskip}     \fi
\ifx \showISSN     \undefined \def \showISSN      #1{\unskip}     \fi
\ifx \showLCCN     \undefined \def \showLCCN      #1{\unskip}     \fi
\ifx \shownote     \undefined \def \shownote      #1{#1}          \fi
\ifx \showarticletitle \undefined \def \showarticletitle #1{#1}   \fi
\ifx \showURL      \undefined \def \showURL       {\relax}        \fi
\providecommand\bibfield[2]{#2}
\providecommand\bibinfo[2]{#2}
\providecommand\natexlab[1]{#1}
\providecommand\showeprint[2][]{arXiv:#2}

\bibitem[Avrahami et~al\mbox{.}(2023)]%
        {avrahami2023break}
\bibfield{author}{\bibinfo{person}{Omri Avrahami}, \bibinfo{person}{Kfir Aberman}, \bibinfo{person}{Ohad Fried}, \bibinfo{person}{Daniel Cohen-Or}, {and} \bibinfo{person}{Dani Lischinski}.} \bibinfo{year}{2023}\natexlab{}.
\newblock \showarticletitle{Break-A-Scene: Extracting Multiple Concepts from a Single Image}.
\newblock \bibinfo{journal}{\emph{arXiv preprint arXiv:2305.16311}} (\bibinfo{year}{2023}).
\newblock


\bibitem[Azadi et~al\mbox{.}(2018)]%
        {azadi2018multi}
\bibfield{author}{\bibinfo{person}{Samaneh Azadi}, \bibinfo{person}{Matthew Fisher}, \bibinfo{person}{Vladimir~G Kim}, \bibinfo{person}{Zhaowen Wang}, \bibinfo{person}{Eli Shechtman}, {and} \bibinfo{person}{Trevor Darrell}.} \bibinfo{year}{2018}\natexlab{}.
\newblock \showarticletitle{Multi-content gan for few-shot font style transfer}. In \bibinfo{booktitle}{\emph{Proceedings of the IEEE conference on computer vision and pattern recognition}}. \bibinfo{pages}{7564--7573}.
\newblock


\bibitem[Bar-Tal et~al\mbox{.}(2023)]%
        {bar2023multidiffusion}
\bibfield{author}{\bibinfo{person}{Omer Bar-Tal}, \bibinfo{person}{Lior Yariv}, \bibinfo{person}{Yaron Lipman}, {and} \bibinfo{person}{Tali Dekel}.} \bibinfo{year}{2023}\natexlab{}.
\newblock \showarticletitle{Multidiffusion: Fusing diffusion paths for controlled image generation}.
\newblock  (\bibinfo{year}{2023}).
\newblock


\bibitem[Barnes et~al\mbox{.}(2009)]%
        {barnes2009patchmatch}
\bibfield{author}{\bibinfo{person}{Connelly Barnes}, \bibinfo{person}{Eli Shechtman}, \bibinfo{person}{Adam Finkelstein}, {and} \bibinfo{person}{Dan~B Goldman}.} \bibinfo{year}{2009}\natexlab{}.
\newblock \showarticletitle{PatchMatch: A randomized correspondence algorithm for structural image editing}.
\newblock \bibinfo{journal}{\emph{ACM Trans. Graph.}} \bibinfo{volume}{28}, \bibinfo{number}{3} (\bibinfo{year}{2009}), \bibinfo{pages}{24}.
\newblock


\bibitem[Chen et~al\mbox{.}(2023)]%
        {chen2023suti}
\bibfield{author}{\bibinfo{person}{Wenhu Chen}, \bibinfo{person}{Hexiang Hu}, \bibinfo{person}{Yandong Li}, \bibinfo{person}{Nataniel Ruiz}, \bibinfo{person}{Xuhui Jia}, \bibinfo{person}{Ming-Wei Chang}, {and} \bibinfo{person}{William~W Cohen}.} \bibinfo{year}{2023}\natexlab{}.
\newblock \showarticletitle{Subject-driven Text-to-Image Generation via Apprenticeship Learning}.
\newblock \bibinfo{journal}{\emph{arXiv preprint arXiv:2304.00186}} (\bibinfo{year}{2023}).
\newblock


\bibitem[Efros and Leung(1999)]%
        {efros1999texture}
\bibfield{author}{\bibinfo{person}{Alexei~A Efros} {and} \bibinfo{person}{Thomas~K Leung}.} \bibinfo{year}{1999}\natexlab{}.
\newblock \showarticletitle{Texture synthesis by non-parametric sampling}. In \bibinfo{booktitle}{\emph{Proceedings of the seventh IEEE international conference on computer vision}}, Vol.~\bibinfo{volume}{2}. IEEE, \bibinfo{pages}{1033--1038}.
\newblock


\bibitem[Face({[n.\,d.]})]%
        {promptweighting}
\bibfield{author}{\bibinfo{person}{Hugging Face}.} \bibinfo{year}{[n.\,d.]}\natexlab{}.
\newblock \bibinfo{title}{Prompt weighting}.
\newblock \bibinfo{howpublished}{\url{https://huggingface.co/docs/diffusers/using-diffusers/weighted_prompts}}.
\newblock


\bibitem[Gal et~al\mbox{.}(2022)]%
        {gal2022image}
\bibfield{author}{\bibinfo{person}{Rinon Gal}, \bibinfo{person}{Yuval Alaluf}, \bibinfo{person}{Yuval Atzmon}, \bibinfo{person}{Or Patashnik}, \bibinfo{person}{Amit~H Bermano}, \bibinfo{person}{Gal Chechik}, {and} \bibinfo{person}{Daniel Cohen-Or}.} \bibinfo{year}{2022}\natexlab{}.
\newblock \showarticletitle{An image is worth one word: Personalizing text-to-image generation using textual inversion}.
\newblock \bibinfo{journal}{\emph{arXiv preprint arXiv:2208.01618}} (\bibinfo{year}{2022}).
\newblock


\bibitem[Gatys et~al\mbox{.}(2015)]%
        {gatys2015texture}
\bibfield{author}{\bibinfo{person}{Leon Gatys}, \bibinfo{person}{Alexander~S Ecker}, {and} \bibinfo{person}{Matthias Bethge}.} \bibinfo{year}{2015}\natexlab{}.
\newblock \showarticletitle{Texture synthesis using convolutional neural networks}.
\newblock \bibinfo{journal}{\emph{Advances in neural information processing systems}}  \bibinfo{volume}{28} (\bibinfo{year}{2015}).
\newblock


\bibitem[Haeberli(1990)]%
        {haeberli1990paint}
\bibfield{author}{\bibinfo{person}{Paul Haeberli}.} \bibinfo{year}{1990}\natexlab{}.
\newblock \showarticletitle{Paint by numbers: Abstract image representations}. In \bibinfo{booktitle}{\emph{Proceedings of the 17th annual conference on Computer graphics and interactive techniques}}. \bibinfo{pages}{207--214}.
\newblock


\bibitem[H{\"a}rk{\"o}nen et~al\mbox{.}(2020)]%
        {harkonen2020ganspace}
\bibfield{author}{\bibinfo{person}{Erik H{\"a}rk{\"o}nen}, \bibinfo{person}{Aaron Hertzmann}, \bibinfo{person}{Jaakko Lehtinen}, {and} \bibinfo{person}{Sylvain Paris}.} \bibinfo{year}{2020}\natexlab{}.
\newblock \showarticletitle{Ganspace: Discovering interpretable gan controls}.
\newblock \bibinfo{journal}{\emph{Advances in neural information processing systems}}  \bibinfo{volume}{33} (\bibinfo{year}{2020}), \bibinfo{pages}{9841--9850}.
\newblock


\bibitem[Heeger and Bergen(1995)]%
        {heeger1995pyramid}
\bibfield{author}{\bibinfo{person}{David~J Heeger} {and} \bibinfo{person}{James~R Bergen}.} \bibinfo{year}{1995}\natexlab{}.
\newblock \showarticletitle{Pyramid-based texture analysis/synthesis}. In \bibinfo{booktitle}{\emph{Proceedings of the 22nd annual conference on Computer graphics and interactive techniques}}. \bibinfo{pages}{229--238}.
\newblock


\bibitem[Hertz et~al\mbox{.}(2022)]%
        {hertz2022prompt}
\bibfield{author}{\bibinfo{person}{Amir Hertz}, \bibinfo{person}{Ron Mokady}, \bibinfo{person}{Jay Tenenbaum}, \bibinfo{person}{Kfir Aberman}, \bibinfo{person}{Yael Pritch}, {and} \bibinfo{person}{Daniel Cohen-Or}.} \bibinfo{year}{2022}\natexlab{}.
\newblock \showarticletitle{Prompt-to-prompt image editing with cross attention control}.
\newblock \bibinfo{journal}{\emph{arXiv preprint arXiv:2208.01626}} (\bibinfo{year}{2022}).
\newblock


\bibitem[Hertzmann(2003)]%
        {hertzmann2003survey}
\bibfield{author}{\bibinfo{person}{Aaron Hertzmann}.} \bibinfo{year}{2003}\natexlab{}.
\newblock \showarticletitle{A survey of stroke-based rendering}. Institute of Electrical and Electronics Engineers.
\newblock


\bibitem[Hertzmann et~al\mbox{.}(2023)]%
        {hertzmann2023image}
\bibfield{author}{\bibinfo{person}{Aaron Hertzmann}, \bibinfo{person}{Charles~E Jacobs}, \bibinfo{person}{Nuria Oliver}, \bibinfo{person}{Brian Curless}, {and} \bibinfo{person}{David~H Salesin}.} \bibinfo{year}{2023}\natexlab{}.
\newblock \showarticletitle{Image analogies}.
\newblock In \bibinfo{booktitle}{\emph{Seminal Graphics Papers: Pushing the Boundaries, Volume 2}}. \bibinfo{pages}{557--570}.
\newblock


\bibitem[Ho et~al\mbox{.}(2020)]%
        {ho2020denoising}
\bibfield{author}{\bibinfo{person}{Jonathan Ho}, \bibinfo{person}{Ajay Jain}, {and} \bibinfo{person}{Pieter Abbeel}.} \bibinfo{year}{2020}\natexlab{}.
\newblock \showarticletitle{Denoising diffusion probabilistic models}.
\newblock \bibinfo{journal}{\emph{Advances in neural information processing systems}}  \bibinfo{volume}{33} (\bibinfo{year}{2020}), \bibinfo{pages}{6840--6851}.
\newblock


\bibitem[Ho and Salimans(2022)]%
        {ho2022classifier}
\bibfield{author}{\bibinfo{person}{Jonathan Ho} {and} \bibinfo{person}{Tim Salimans}.} \bibinfo{year}{2022}\natexlab{}.
\newblock \showarticletitle{Classifier-free diffusion guidance}.
\newblock \bibinfo{journal}{\emph{arXiv preprint arXiv:2207.12598}} (\bibinfo{year}{2022}).
\newblock


\bibitem[Houlsby et~al\mbox{.}(2019)]%
        {houlsby2019parameter}
\bibfield{author}{\bibinfo{person}{Neil Houlsby}, \bibinfo{person}{Andrei Giurgiu}, \bibinfo{person}{Stanislaw Jastrzebski}, \bibinfo{person}{Bruna Morrone}, \bibinfo{person}{Quentin De~Laroussilhe}, \bibinfo{person}{Andrea Gesmundo}, \bibinfo{person}{Mona Attariyan}, {and} \bibinfo{person}{Sylvain Gelly}.} \bibinfo{year}{2019}\natexlab{}.
\newblock \showarticletitle{Parameter-efficient transfer learning for NLP}. In \bibinfo{booktitle}{\emph{International Conference on Machine Learning}}. PMLR, \bibinfo{pages}{2790--2799}.
\newblock


\bibitem[Hu et~al\mbox{.}(2021)]%
        {hu2021lora}
\bibfield{author}{\bibinfo{person}{Edward~J Hu}, \bibinfo{person}{Yelong Shen}, \bibinfo{person}{Phillip Wallis}, \bibinfo{person}{Zeyuan Allen-Zhu}, \bibinfo{person}{Yuanzhi Li}, \bibinfo{person}{Shean Wang}, \bibinfo{person}{Lu Wang}, {and} \bibinfo{person}{Weizhu Chen}.} \bibinfo{year}{2021}\natexlab{}.
\newblock \showarticletitle{Lora: Low-rank adaptation of large language models}.
\newblock \bibinfo{journal}{\emph{arXiv preprint arXiv:2106.09685}} (\bibinfo{year}{2021}).
\newblock


\bibitem[Jahanian et~al\mbox{.}(2019)]%
        {jahanian2019steerability}
\bibfield{author}{\bibinfo{person}{Ali Jahanian}, \bibinfo{person}{Lucy Chai}, {and} \bibinfo{person}{Phillip Isola}.} \bibinfo{year}{2019}\natexlab{}.
\newblock \showarticletitle{On the" steerability" of generative adversarial networks}.
\newblock \bibinfo{journal}{\emph{arXiv preprint arXiv:1907.07171}} (\bibinfo{year}{2019}).
\newblock


\bibitem[Jing et~al\mbox{.}(2019)]%
        {jing2019neural}
\bibfield{author}{\bibinfo{person}{Yongcheng Jing}, \bibinfo{person}{Yezhou Yang}, \bibinfo{person}{Zunlei Feng}, \bibinfo{person}{Jingwen Ye}, \bibinfo{person}{Yizhou Yu}, {and} \bibinfo{person}{Mingli Song}.} \bibinfo{year}{2019}\natexlab{}.
\newblock \showarticletitle{Neural style transfer: A review}.
\newblock \bibinfo{journal}{\emph{IEEE transactions on visualization and computer graphics}} \bibinfo{volume}{26}, \bibinfo{number}{11} (\bibinfo{year}{2019}), \bibinfo{pages}{3365--3385}.
\newblock


\bibitem[Karras et~al\mbox{.}(2019)]%
        {kerras2019style}
\bibfield{author}{\bibinfo{person}{Tero Karras}, \bibinfo{person}{Samuli Laine}, {and} \bibinfo{person}{Timo Aila}.} \bibinfo{year}{2019}\natexlab{}.
\newblock \showarticletitle{A Style-Based Generator Architecture for Generative Adversarial Networks}. In \bibinfo{booktitle}{\emph{2019 IEEE/CVF Conference on Computer Vision and Pattern Recognition (CVPR)}}. \bibinfo{pages}{4396--4405}.
\newblock
\urldef\tempurl%
\url{https://doi.org/10.1109/CVPR.2019.00453}
\showDOI{\tempurl}


\bibitem[Kumari et~al\mbox{.}(2023)]%
        {kumari2023multi}
\bibfield{author}{\bibinfo{person}{Nupur Kumari}, \bibinfo{person}{Bingliang Zhang}, \bibinfo{person}{Richard Zhang}, \bibinfo{person}{Eli Shechtman}, {and} \bibinfo{person}{Jun-Yan Zhu}.} \bibinfo{year}{2023}\natexlab{}.
\newblock \showarticletitle{Multi-concept customization of text-to-image diffusion}. In \bibinfo{booktitle}{\emph{Proceedings of the IEEE/CVF Conference on Computer Vision and Pattern Recognition}}. \bibinfo{pages}{1931--1941}.
\newblock


\bibitem[Lee et~al\mbox{.}(2023)]%
        {lee2023syncdiffusion}
\bibfield{author}{\bibinfo{person}{Yuseung Lee}, \bibinfo{person}{Kunho Kim}, \bibinfo{person}{Hyunjin Kim}, {and} \bibinfo{person}{Minhyuk Sung}.} \bibinfo{year}{2023}\natexlab{}.
\newblock \showarticletitle{Syncdiffusion: Coherent montage via synchronized joint diffusions}.
\newblock \bibinfo{journal}{\emph{arXiv preprint arXiv:2306.05178}} (\bibinfo{year}{2023}).
\newblock


\bibitem[Li et~al\mbox{.}(2023)]%
        {li2023gligen}
\bibfield{author}{\bibinfo{person}{Yuheng Li}, \bibinfo{person}{Haotian Liu}, \bibinfo{person}{Qingyang Wu}, \bibinfo{person}{Fangzhou Mu}, \bibinfo{person}{Jianwei Yang}, \bibinfo{person}{Jianfeng Gao}, \bibinfo{person}{Chunyuan Li}, {and} \bibinfo{person}{Yong~Jae Lee}.} \bibinfo{year}{2023}\natexlab{}.
\newblock \showarticletitle{Gligen: Open-set grounded text-to-image generation}. In \bibinfo{booktitle}{\emph{Proceedings of the IEEE/CVF Conference on Computer Vision and Pattern Recognition}}. \bibinfo{pages}{22511--22521}.
\newblock


\bibitem[Litwinowicz(1997)]%
        {litwinowicz1997processing}
\bibfield{author}{\bibinfo{person}{Peter Litwinowicz}.} \bibinfo{year}{1997}\natexlab{}.
\newblock \showarticletitle{Processing images and video for an impressionist effect}. In \bibinfo{booktitle}{\emph{Proceedings of the 24th annual conference on Computer graphics and interactive techniques}}. \bibinfo{pages}{407--414}.
\newblock


\bibitem[Liu et~al\mbox{.}(2022)]%
        {liu2022compositional}
\bibfield{author}{\bibinfo{person}{Nan Liu}, \bibinfo{person}{Shuang Li}, \bibinfo{person}{Yilun Du}, \bibinfo{person}{Antonio Torralba}, {and} \bibinfo{person}{Joshua~B Tenenbaum}.} \bibinfo{year}{2022}\natexlab{}.
\newblock \showarticletitle{Compositional visual generation with composable diffusion models}. In \bibinfo{booktitle}{\emph{European Conference on Computer Vision}}. Springer, \bibinfo{pages}{423--439}.
\newblock


\bibitem[Ma et~al\mbox{.}(2023)]%
        {ma2023subject}
\bibfield{author}{\bibinfo{person}{Jian Ma}, \bibinfo{person}{Junhao Liang}, \bibinfo{person}{Chen Chen}, {and} \bibinfo{person}{Haonan Lu}.} \bibinfo{year}{2023}\natexlab{}.
\newblock \showarticletitle{Subject-diffusion: Open domain personalized text-to-image generation without test-time fine-tuning}.
\newblock \bibinfo{journal}{\emph{arXiv preprint arXiv:2307.11410}} (\bibinfo{year}{2023}).
\newblock


\bibitem[Meng et~al\mbox{.}(2021)]%
        {meng2021sdedit}
\bibfield{author}{\bibinfo{person}{Chenlin Meng}, \bibinfo{person}{Yutong He}, \bibinfo{person}{Yang Song}, \bibinfo{person}{Jiaming Song}, \bibinfo{person}{Jiajun Wu}, \bibinfo{person}{Jun-Yan Zhu}, {and} \bibinfo{person}{Stefano Ermon}.} \bibinfo{year}{2021}\natexlab{}.
\newblock \showarticletitle{Sdedit: Guided image synthesis and editing with stochastic differential equations}.
\newblock \bibinfo{journal}{\emph{arXiv preprint arXiv:2108.01073}} (\bibinfo{year}{2021}).
\newblock


\bibitem[Park et~al\mbox{.}(2023)]%
        {park2023understanding}
\bibfield{author}{\bibinfo{person}{Yong-Hyun Park}, \bibinfo{person}{Mingi Kwon}, \bibinfo{person}{Jaewoong Choi}, \bibinfo{person}{Junghyo Jo}, {and} \bibinfo{person}{Youngjung Uh}.} \bibinfo{year}{2023}\natexlab{}.
\newblock \showarticletitle{Understanding the latent space of diffusion models through the lens of riemannian geometry}.
\newblock \bibinfo{journal}{\emph{arXiv preprint arXiv:2307.12868}} (\bibinfo{year}{2023}).
\newblock


\bibitem[Patashnik et~al\mbox{.}(2021)]%
        {patashnik2021styleclip}
\bibfield{author}{\bibinfo{person}{Or Patashnik}, \bibinfo{person}{Zongze Wu}, \bibinfo{person}{Eli Shechtman}, \bibinfo{person}{Daniel Cohen-Or}, {and} \bibinfo{person}{Dani Lischinski}.} \bibinfo{year}{2021}\natexlab{}.
\newblock \showarticletitle{Styleclip: Text-driven manipulation of stylegan imagery}. In \bibinfo{booktitle}{\emph{Proceedings of the IEEE/CVF International Conference on Computer Vision}}. \bibinfo{pages}{2085--2094}.
\newblock


\bibitem[Portilla and Simoncelli(2000)]%
        {portilla2000parametric}
\bibfield{author}{\bibinfo{person}{Javier Portilla} {and} \bibinfo{person}{Eero~P Simoncelli}.} \bibinfo{year}{2000}\natexlab{}.
\newblock \showarticletitle{A parametric texture model based on joint statistics of complex wavelet coefficients}.
\newblock \bibinfo{journal}{\emph{International journal of computer vision}}  \bibinfo{volume}{40} (\bibinfo{year}{2000}), \bibinfo{pages}{49--70}.
\newblock


\bibitem[Preechakul et~al\mbox{.}(2022)]%
        {preechakul2022diffusion}
\bibfield{author}{\bibinfo{person}{Konpat Preechakul}, \bibinfo{person}{Nattanat Chatthee}, \bibinfo{person}{Suttisak Wizadwongsa}, {and} \bibinfo{person}{Supasorn Suwajanakorn}.} \bibinfo{year}{2022}\natexlab{}.
\newblock \showarticletitle{Diffusion autoencoders: Toward a meaningful and decodable representation}. In \bibinfo{booktitle}{\emph{Proceedings of the IEEE/CVF Conference on Computer Vision and Pattern Recognition}}. \bibinfo{pages}{10619--10629}.
\newblock


\bibitem[Rashtchian et~al\mbox{.}(2023)]%
        {rashtchian2023substance}
\bibfield{author}{\bibinfo{person}{Cyrus Rashtchian}, \bibinfo{person}{Charles Herrmann}, \bibinfo{person}{Chun-Sung Ferng}, \bibinfo{person}{Ayan Chakrabarti}, \bibinfo{person}{Dilip Krishnan}, \bibinfo{person}{Deqing Sun}, \bibinfo{person}{Da-Cheng Juan}, {and} \bibinfo{person}{Andrew Tomkins}.} \bibinfo{year}{2023}\natexlab{}.
\newblock \showarticletitle{Substance or Style: What Does Your Image Embedding Know?}
\newblock \bibinfo{journal}{\emph{arXiv preprint arXiv:2307.05610}} (\bibinfo{year}{2023}).
\newblock


\bibitem[Rissanen et~al\mbox{.}(2022)]%
        {rissanen2022generative}
\bibfield{author}{\bibinfo{person}{Severi Rissanen}, \bibinfo{person}{Markus Heinonen}, {and} \bibinfo{person}{Arno Solin}.} \bibinfo{year}{2022}\natexlab{}.
\newblock \showarticletitle{Generative modelling with inverse heat dissipation}.
\newblock \bibinfo{journal}{\emph{arXiv preprint arXiv:2206.13397}} (\bibinfo{year}{2022}).
\newblock


\bibitem[Rombach et~al\mbox{.}(2022)]%
        {rombach2022high}
\bibfield{author}{\bibinfo{person}{Robin Rombach}, \bibinfo{person}{Andreas Blattmann}, \bibinfo{person}{Dominik Lorenz}, \bibinfo{person}{Patrick Esser}, {and} \bibinfo{person}{Bj{\"o}rn Ommer}.} \bibinfo{year}{2022}\natexlab{}.
\newblock \showarticletitle{High-resolution image synthesis with latent diffusion models}. In \bibinfo{booktitle}{\emph{Proceedings of the IEEE/CVF conference on computer vision and pattern recognition}}. \bibinfo{pages}{10684--10695}.
\newblock


\bibitem[Ruiz et~al\mbox{.}(2023)]%
        {ruiz2023dreambooth}
\bibfield{author}{\bibinfo{person}{Nataniel Ruiz}, \bibinfo{person}{Yuanzhen Li}, \bibinfo{person}{Varun Jampani}, \bibinfo{person}{Yael Pritch}, \bibinfo{person}{Michael Rubinstein}, {and} \bibinfo{person}{Kfir Aberman}.} \bibinfo{year}{2023}\natexlab{}.
\newblock \showarticletitle{Dreambooth: Fine tuning text-to-image diffusion models for subject-driven generation}. In \bibinfo{booktitle}{\emph{Proceedings of the IEEE/CVF Conference on Computer Vision and Pattern Recognition}}. \bibinfo{pages}{22500--22510}.
\newblock


\bibitem[Saharia et~al\mbox{.}(2022)]%
        {saharia2022photorealistic}
\bibfield{author}{\bibinfo{person}{Chitwan Saharia}, \bibinfo{person}{William Chan}, \bibinfo{person}{Saurabh Saxena}, \bibinfo{person}{Lala Li}, \bibinfo{person}{Jay Whang}, \bibinfo{person}{Emily~L Denton}, \bibinfo{person}{Kamyar Ghasemipour}, \bibinfo{person}{Raphael Gontijo~Lopes}, \bibinfo{person}{Burcu Karagol~Ayan}, \bibinfo{person}{Tim Salimans}, {et~al\mbox{.}}} \bibinfo{year}{2022}\natexlab{}.
\newblock \showarticletitle{Photorealistic text-to-image diffusion models with deep language understanding}.
\newblock \bibinfo{journal}{\emph{Advances in Neural Information Processing Systems}}  \bibinfo{volume}{35} (\bibinfo{year}{2022}), \bibinfo{pages}{36479--36494}.
\newblock


\bibitem[Salisbury et~al\mbox{.}(2023)]%
        {salisbury2023interactive}
\bibfield{author}{\bibinfo{person}{Michaet~P Salisbury}, \bibinfo{person}{Sean~E Anderson}, \bibinfo{person}{Ronen Barzel}, {and} \bibinfo{person}{David~H Satesin}.} \bibinfo{year}{2023}\natexlab{}.
\newblock \showarticletitle{Interactive pen-and-ink illustration}.
\newblock In \bibinfo{booktitle}{\emph{Seminal Graphics Papers: Pushing the Boundaries, Volume 2}}. \bibinfo{pages}{393--400}.
\newblock


\bibitem[Shen et~al\mbox{.}(2020)]%
        {shen2020interpreting}
\bibfield{author}{\bibinfo{person}{Yujun Shen}, \bibinfo{person}{Jinjin Gu}, \bibinfo{person}{Xiaoou Tang}, {and} \bibinfo{person}{Bolei Zhou}.} \bibinfo{year}{2020}\natexlab{}.
\newblock \showarticletitle{Interpreting the latent space of gans for semantic face editing}. In \bibinfo{booktitle}{\emph{Proceedings of the IEEE/CVF conference on computer vision and pattern recognition}}. \bibinfo{pages}{9243--9252}.
\newblock


\bibitem[Shi et~al\mbox{.}(2023)]%
        {shi2023instantbooth}
\bibfield{author}{\bibinfo{person}{Jing Shi}, \bibinfo{person}{Wei Xiong}, \bibinfo{person}{Zhe Lin}, {and} \bibinfo{person}{Hyun~Joon Jung}.} \bibinfo{year}{2023}\natexlab{}.
\newblock \showarticletitle{Instantbooth: Personalized text-to-image generation without test-time finetuning}.
\newblock \bibinfo{journal}{\emph{arXiv preprint arXiv:2304.03411}} (\bibinfo{year}{2023}).
\newblock


\bibitem[Sohl-Dickstein et~al\mbox{.}(2015)]%
        {sohl2015deep}
\bibfield{author}{\bibinfo{person}{Jascha Sohl-Dickstein}, \bibinfo{person}{Eric Weiss}, \bibinfo{person}{Niru Maheswaranathan}, {and} \bibinfo{person}{Surya Ganguli}.} \bibinfo{year}{2015}\natexlab{}.
\newblock \showarticletitle{Deep unsupervised learning using nonequilibrium thermodynamics}. In \bibinfo{booktitle}{\emph{International conference on machine learning}}. PMLR, \bibinfo{pages}{2256--2265}.
\newblock


\bibitem[Sohn et~al\mbox{.}(2023)]%
        {sohn2023styledrop}
\bibfield{author}{\bibinfo{person}{Kihyuk Sohn}, \bibinfo{person}{Nataniel Ruiz}, \bibinfo{person}{Kimin Lee}, \bibinfo{person}{Daniel~Castro Chin}, \bibinfo{person}{Irina Blok}, \bibinfo{person}{Huiwen Chang}, \bibinfo{person}{Jarred Barber}, \bibinfo{person}{Lu Jiang}, \bibinfo{person}{Glenn Entis}, \bibinfo{person}{Yuanzhen Li}, {et~al\mbox{.}}} \bibinfo{year}{2023}\natexlab{}.
\newblock \showarticletitle{StyleDrop: Text-to-Image Generation in Any Style}.
\newblock \bibinfo{journal}{\emph{arXiv preprint arXiv:2306.00983}} (\bibinfo{year}{2023}).
\newblock


\bibitem[Stewart(2023)]%
        {Compel:GitHub}
\bibfield{author}{\bibinfo{person}{Damian Stewart}.} \bibinfo{year}{2023}\natexlab{}.
\newblock \bibinfo{title}{Compel}.
\newblock \bibinfo{howpublished}{\url{https://github.com/damian0815/compel}}.
\newblock


\bibitem[Tewari et~al\mbox{.}(2020)]%
        {tewari2020stylerig}
\bibfield{author}{\bibinfo{person}{Ayush Tewari}, \bibinfo{person}{Mohamed Elgharib}, \bibinfo{person}{Gaurav Bharaj}, \bibinfo{person}{Florian Bernard}, \bibinfo{person}{Hans-Peter Seidel}, \bibinfo{person}{Patrick P{\'e}rez}, \bibinfo{person}{Michael Zollhofer}, {and} \bibinfo{person}{Christian Theobalt}.} \bibinfo{year}{2020}\natexlab{}.
\newblock \showarticletitle{Stylerig: Rigging stylegan for 3d control over portrait images}. In \bibinfo{booktitle}{\emph{Proceedings of the IEEE/CVF Conference on Computer Vision and Pattern Recognition}}. \bibinfo{pages}{6142--6151}.
\newblock


\bibitem[Voynov and Babenko(2020)]%
        {voynov2020unsupervised}
\bibfield{author}{\bibinfo{person}{Andrey Voynov} {and} \bibinfo{person}{Artem Babenko}.} \bibinfo{year}{2020}\natexlab{}.
\newblock \showarticletitle{Unsupervised discovery of interpretable directions in the gan latent space}. In \bibinfo{booktitle}{\emph{International conference on machine learning}}. PMLR, \bibinfo{pages}{9786--9796}.
\newblock


\bibitem[Wu et~al\mbox{.}(2023a)]%
        {wu2023uncovering}
\bibfield{author}{\bibinfo{person}{Qiucheng Wu}, \bibinfo{person}{Yujian Liu}, \bibinfo{person}{Handong Zhao}, \bibinfo{person}{Ajinkya Kale}, \bibinfo{person}{Trung Bui}, \bibinfo{person}{Tong Yu}, \bibinfo{person}{Zhe Lin}, \bibinfo{person}{Yang Zhang}, {and} \bibinfo{person}{Shiyu Chang}.} \bibinfo{year}{2023}\natexlab{a}.
\newblock \showarticletitle{Uncovering the disentanglement capability in text-to-image diffusion models}. In \bibinfo{booktitle}{\emph{Proceedings of the IEEE/CVF Conference on Computer Vision and Pattern Recognition}}. \bibinfo{pages}{1900--1910}.
\newblock


\bibitem[Wu et~al\mbox{.}(2023b)]%
        {wu2023not}
\bibfield{author}{\bibinfo{person}{Yankun Wu}, \bibinfo{person}{Yuta Nakashima}, {and} \bibinfo{person}{Noa Garcia}.} \bibinfo{year}{2023}\natexlab{b}.
\newblock \showarticletitle{Not Only Generative Art: Stable Diffusion for Content-Style Disentanglement in Art Analysis}. In \bibinfo{booktitle}{\emph{Proceedings of the 2023 ACM International Conference on Multimedia Retrieval}}. \bibinfo{pages}{199--208}.
\newblock


\bibitem[Wu et~al\mbox{.}(2021)]%
        {wu2021stylespace}
\bibfield{author}{\bibinfo{person}{Zongze Wu}, \bibinfo{person}{Dani Lischinski}, {and} \bibinfo{person}{Eli Shechtman}.} \bibinfo{year}{2021}\natexlab{}.
\newblock \showarticletitle{Stylespace analysis: Disentangled controls for stylegan image generation}. In \bibinfo{booktitle}{\emph{Proceedings of the IEEE/CVF Conference on Computer Vision and Pattern Recognition}}. \bibinfo{pages}{12863--12872}.
\newblock


\bibitem[Yang et~al\mbox{.}(2022)]%
        {Yang2022DiffusionMA}
\bibfield{author}{\bibinfo{person}{Ling Yang}, \bibinfo{person}{Zhilong Zhang}, \bibinfo{person}{Yang Song}, \bibinfo{person}{Shenda Hong}, \bibinfo{person}{Runsheng Xu}, \bibinfo{person}{Yue Zhao}, \bibinfo{person}{Yingxia Shao}, \bibinfo{person}{Wentao Zhang}, \bibinfo{person}{Bin Cui}, {and} \bibinfo{person}{Ming-Hsuan Yang}.} \bibinfo{year}{2022}\natexlab{}.
\newblock \showarticletitle{Diffusion models: A comprehensive survey of methods and applications}.
\newblock \bibinfo{journal}{\emph{arXiv preprint arXiv:2209.00796}} (\bibinfo{year}{2022}).
\newblock


\bibitem[Zhang et~al\mbox{.}(2023)]%
        {zhang2023adding}
\bibfield{author}{\bibinfo{person}{Lvmin Zhang}, \bibinfo{person}{Anyi Rao}, {and} \bibinfo{person}{Maneesh Agrawala}.} \bibinfo{year}{2023}\natexlab{}.
\newblock \bibinfo{title}{Adding Conditional Control to Text-to-Image Diffusion Models}.
\newblock
\newblock


\end{thebibliography}
}

\end{document}